%% file: main.tex
\documentclass[10pt,twocolumn,letterpaper]{article}

\usepackage[pagenumbers]{cvpr} 

\usepackage{graphicx}
\usepackage{amsmath}
\usepackage{amssymb}
\usepackage{booktabs}

\usepackage{url}
\usepackage{epsfig}
\usepackage[accsupp]{axessibility}
\usepackage{helvet}   
\usepackage{courier}  
\usepackage{url}      
\usepackage{xcolor}
\usepackage{multirow}
\usepackage{subcaption}
\usepackage{algorithm}
\usepackage{algpseudocode}
\usepackage[export]{adjustbox}
\usepackage{color}
\usepackage{pifont}
\definecolor{mycolor}{RGB}{147,112,219}
\definecolor{yccoloer}{RGB}{52, 235, 150}
\usepackage{appendix}
\usepackage{amsmath}

\usepackage[pagebackref,breaklinks,colorlinks]{hyperref}

\usepackage[capitalize]{cleveref}
\crefname{section}{Sec.}{Secs.}
\Crefname{section}{Section}{Sections}
\Crefname{table}{Table}{Tables}
\crefname{table}{Tab.}{Tabs.}

\def\cvprPaperID{2132} 
\def\confName{CVPR}
\def\confYear{2023}

\begin{document}

\title{Learning an Encoder for 3D Aware Generative NeRF model to Render Object from a Single Image}
\title{Learning an Encoder for Rendering Single Image\\ with 3D-Aware Generative NeRF Model}
\title{Learning an Encoder for Inversion of Single Image\\ with 3D-Aware Generative Radiance Field}
\title{Learning an Encoder for GAN Inversion of 3D-Aware Generative Radiance Field}
\title{Learning an Encoder for Inversion of 3D-Aware Style-based Radiance Field}
\title{Learning a 3D-Aware StyleNeRF Encoder for Single-to-Multiview image synthesis}
\title{Learning a 3D-Aware Style-based NeRF Encoder for 2D-to-3D image translation}
\title{Learning a 3D-Aware Encoder for Style-based Generative Radiance Field}
\title{3D-Aware Encoding for Style-based Neural Radiance Fields}

\author{
Yu-Jhe Li$^{1,2}$\thanks{~Work done during an internship at Meta Research.} \qquad Tao Xu$^2$ \qquad Bichen Wu$^2$ \qquad Ningyuan Zheng$^2$\qquad Xiaoliang Dai$^2$\\
Albert Pumarola$^2$ \qquad Peizhao Zhang$^2$ \qquad Peter Vajda$^2$ \qquad Kris Kitani$^1$
\\
$^{1}$Carnegie Mellon University
\qquad\qquad
$^{2}$Meta Research\\
{\tt\small \{\url{yujheli},\url{kkitani}\}\url{@cs.cmu.edu}} \\
{\tt\small \{\url{xutao},\url{wbc},\url{zhengningyuan},\url{xiaoliangdai},\url{apumarola},\url{stzpz},\url{vajdap}\}\url{@meta.com}} \\
{\tt\small }
\vspace{-10mm}
}
\maketitle
\input{0-abstract}
\input{1-introduction}

\input{2-related-works}
\input{3-method}

\input{4-experiment}

\input{5-conclusion}

{\small
\bibliographystyle{ieee_fullname}
\bibliography{egbib}
}
\clearpage
\appendix
\input{appendix.tex}
\end{document}

%% file: 0-abstract.tex
\begin{abstract}

We tackle the task of NeRF inversion for style-based neural radiance fields, (\textit{e.g.}, StyleNeRF). In particular, we aim to learn an inversion function to project an input image to the latent space of a NeRF generator and then synthesize novel views based on the latent code. Compared with GAN inversion for 2D generative models, NeRF inversion needs to (1) preserve the identity of the input image and (2) ensure 3D consistency in generated novel views. This requires the latent code obtained from the single-view image to be invariant across multiple views. To address this new challenge, we propose a two-stage encoder for style-based NeRF inversion. In the first stage, we introduce a base encoder that converts the input image to a latent code. To ensure the latent code is view-invariant and is able to synthesize 3D consistent novel view images, we utilize identity contrastive learning to train the base encoder. Second, to better preserve the identity of the input image, we introduce a refining encoder to refine the latent code and add finer details to the output image. Importantly note that the novelty of this model lies in the design of its first-stage encoder which produces the closest latent code lying on the latent manifold, making the refinement in the second stage to be close to the NeRF manifold. Through extensive experiments, we demonstrate that our proposed two-stage encoder qualitatively and quantitatively exhibits superiority over the existing encoders for inversion in both image reconstruction and novel-view rendering.
\end{abstract}

%% file: 1-introduction.tex
\section{Introduction}



\begin{figure}[t!]
  \centering
  \includegraphics[width=\linewidth]{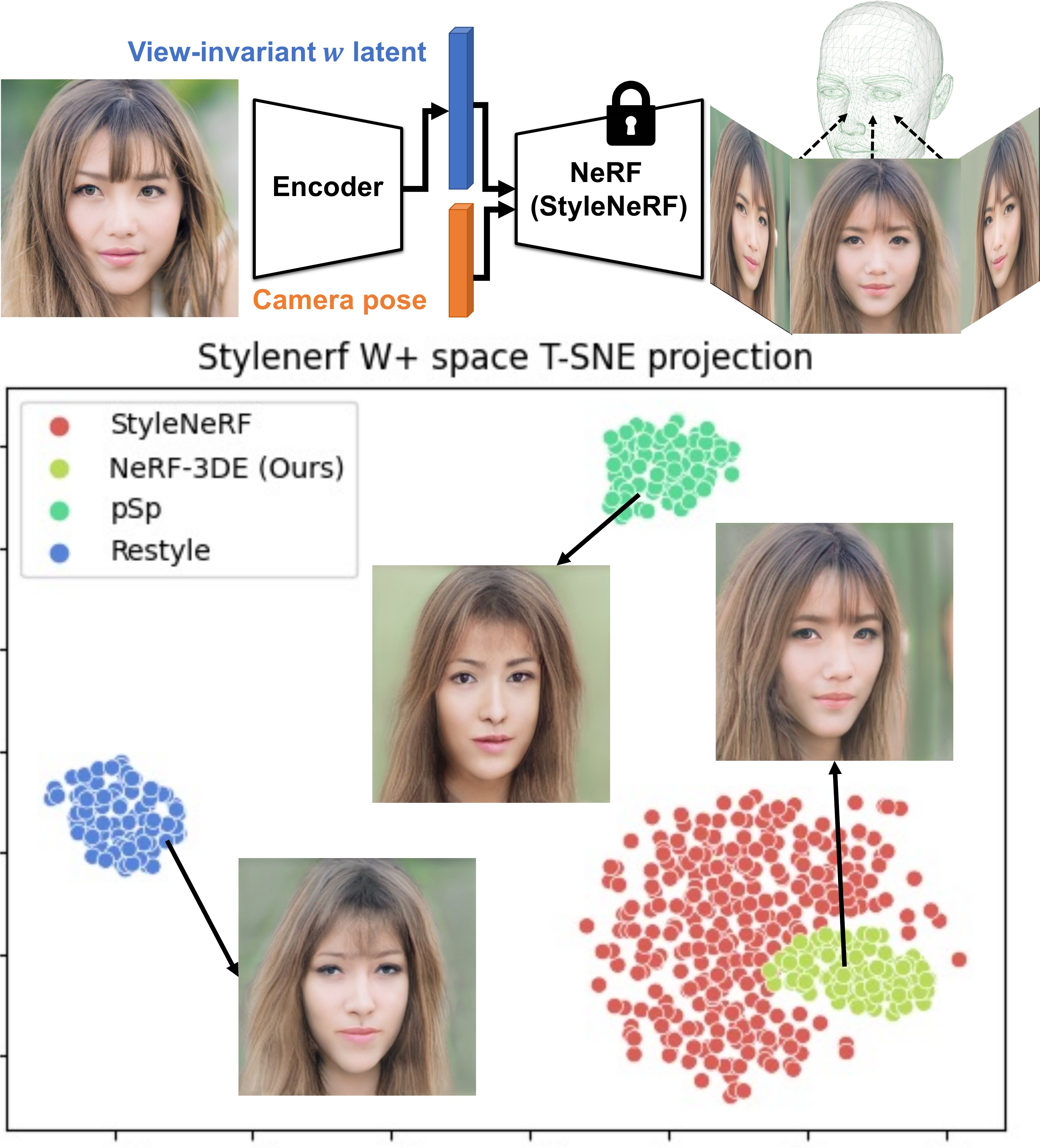}
  \vspace{-6mm}
  \caption{\textbf{Comparison of encoder-based inversion methods for the inversion of style-based NeRF}. Compared with existing effective inversion methods for StyleGAN, it is more challenging to perform inversion on style-based NeRF since $w$ latent is assumed to be view-invariant. Directly using the existing encoder-based methods may not learn this property and may produce the latent lying outside the manifold.}
  \label{fig:teasor}
  \vspace{-4mm}
\end{figure}


We aim at tackling NeRF inversion of 3D style-based generative radiance fields, which typically combine neural radiance field (NeRF)~\cite{mildenhall2020nerf} with the generative adversarial network (GAN)~\cite{goodfellow2014generative}. NeRF inversion is similar to GAN inversion (\cite{zhu2016generative}) which learns a mapping function to project an image into the GAN's latent space. Currently, GAN inversion has been successfully explored in StyleGANs~\cite{karras2019style,Karras2019stylegan2} (\textit{e.g.,} StyleGANv2) which has been used for image synthesis, and enables flexible control of the latent space. Several approaches of GAN inversion are capable of inverting the input image into the latent space (\textit{i.e.}, $\mathcal{W}$ space)~\cite{jahanian2019steerability,shen2020interpreting,tewari2020stylerig,harkonen2020ganspace} or extended latent space (\textit{i.e.}, $\mathcal{W+}$ space: concatenation of all $\mathcal{W}$ latent code from each layer)~\cite{abdal2019image2stylegan,abdal2020image2stylegan++,zhu2020indomain,abdal2021styleflow} for image editing. 
Recently, IDE-3D~\cite{sun2022ide} directly employ a basic $\mathcal{W+}$ encoder for the inversion of their proposed 3D neural renderer with semantic masks. 
However, the exploration of generalized inversion approaches for style-based NeRFs is still limited. 


Recently, 3D style-based models using radiance fields (\textit{i.e.}, NeRF~\cite{mildenhall2020nerf}), such as EG3D~\cite{chan2022efficient} or StyleNeRF~\cite{gu2021stylenerf}, have been proposed for unsupervised generation of multi-view consistent images. Similar to StyleGANs, these NeRFs learn a controllable $\mathcal{W}$ space and enable explicit 3D camera control, using only single-view training images. 
To achieve NeRF inversion for these models, one straightforward way is to directly apply the aforementioned GAN inversion methods, by feeding the image and the corresponding camera poses as inputs and rendering the corresponding multi-view images (see Figure~\ref{fig:teasor}).
However, there are two main challenges. First, if only single-view images are used to train the inversion method, the predicted latent code
only works when generating images of the same view (camera pose), but fails to generate the high-quality image for novel views. This usually happens because the predicted latent is far from the valid latent space and may not be view-invariant, shown in Figure \ref{fig:teasor}. To address this issue, we need multi-view images to train the inversion method. However, this leads to our second challenge: it may be difficult, if not infeasible, to collect sufficient multi-view images with known camera poses for training. 

To learn an inversion function for producing the view-invariant latent code without the use of multi-view images, we propose a framework named NeRF-3DE to learn a 3D-aware Encoder for style-based NeRFs. NeRF-3DE is composed of a two-stage learnable encoder: a base encoder and a refining encoder. 
First, we introduce a base encoder to learn the view-invariant latent code in $\mathcal{W}$ space.
Moreover, we leverage synthesized images (\textit{i.e.}, using the multi-view images generated by the model itself) and contrastive learning with the triplet loss to learn a better view-invariant latent code in $\mathcal{W}$. 
Second, since the latent code in $\mathcal{W}$ space is known to be more difficult to fully reconstruct the input image compared with $\mathcal{W+}$~\cite{Karras2019stylegan2}, we propose a refining encoder to refine the latent code from the base encoder in $\mathcal{W+}$ space. It adds more fine-grained details to the generated image, which makes it more consistent with the input image. Importantly note that the novelty of this model lies in the design of the first-stage base encoder which produces the closest latent code on the latent manifold and thus the refinement in the second stage would still be close to the NeRF latent space.

We have verified the effectiveness of the proposed method and its key components using StyleNeRF~\cite{gu2021stylenerf} as the pre-trained generator for NeRF inversion. Moreover, to test the generalization ability of the proposed method, we combine it with the online optimization technique PTI~\cite{roich2021pivotal}, apply it to a different pretrained generator EG3D~\cite{chan2022efficient}, or utilize CLIP~\cite{radford2021learning} to further edit the produced latent (see supplementary for details).
The contributions of this paper are summarized as follows:


\vspace{-3mm}
\begin{itemize}
\itemsep -1mm
\item We demonstrate the challenges of inversion for style-based NeRFs and the limitations of the current encoder-based models for this task.
\item We propose an encoder-based framework named NeRF-3DE, which aims to produce view-invariant latent for inversion of style-based neural radiance field.
\item Compared with the existing encoders for GAN inversion, our proposed two-stage models achieve more effective inversion for the style-based NeRF and have superior image quality for novel-view rendering.
\item The view-invariant latent produced by our NeRF-3DE has a good generalization to enable online optimization methods (\textit{e.g.} PTI) for rendering novel views and to invert more style-based NeRFs (\textit{e.g.} EG3D).

\end{itemize}
%




%% file: 2-related-works.tex
\section{Related Works}



\paragraph{Latent manipulation in GAN.}

GANs~\cite{goodfellow2014generative} have demonstrated success in image synthesis and have been extended to a number of works~\cite{zhang2019self,brock2018large,karras2017progressive}.
StyleGANs~\cite{karras2019style,Karras2019stylegan2} achieve
state-of-the-art image quality and support different levels of semantic manipulation.
%
%
%
In particular, many methods have been proposed for finding these semantic latent space manipulation using varying levels of supervision. These include full-supervision in the form of semantic
labels~\cite{abdal2021styleflow,shen2020interpreting,goetschalckx2019ganalyze} and unsupervised approaches~\cite{wang2021geometry,voynov2020unsupervised}. Some methods~\cite{harkonen2020ganspace,tewari2020stylerig,abdal2020image2stylegan++,shoshan2021gan} also leverage disentangled properties in the latent space to enable 3D controls. However, most of these works focus on the rendering of 2D images with 3D controls
and are not capable of manipulating camera poses easily as volumetric rendering (NeRF~\cite{mildenhall2020nerf}). 

\paragraph{Image Synthesis with Generative NeRF.}

Methods built on implicit
functions, e.g., NeRF~\cite{mildenhall2020nerf}, have been proposed in \cite{chan2021pi,schwarz2020graf,pan2021shading,niemeyer2021giraffe}.
To generate high-resolution images conditioned on the input style latent code, EG3D~\cite{chan2022efficient}, StyleNeRF~\cite{gu2021stylenerf}, VolumeGAN~\cite{xu20223d}, StyleSDF~\cite{or2022stylesdf}, and GMPI~\cite{zhao2022generative} have been developed. In addition, some works such as Sofgan~\cite{chen2022sofgan} and Sem2NeRF~\cite{chen2022sem2nerf} are able to perform multi-view synthesis with NeRF by taking into multi-view or single-view semantic masks.
Among these models, StyleNeRF~\cite{gu2021stylenerf} is able to perform novel-view image synthesis given the style latent code and the camera pose and only relies on MLP layers as the classical NeRF~\cite{mildenhall2020nerf}. To simplify the analysis of NeRF inversion for style-based NeRF, we employ StyleNeRF in our experiments.   

\paragraph{Inversion.}

GAN inversion~\cite{zhu2016generative} is the process of obtaining a latent code that can allow the generator to reconstruct the given image. Generally, inversion methods either directly optimize the latent feature to minimize the loss for a given image~\cite{abdal2019image2stylegan,abdal2020image2stylegan++,bau2020semantic,gu2020image}, train an encoder on a large number of images to learn a mapping from an image to a style latent~\cite{alaluf2021restyle,guan2020collaborative,kang2021gan,kim2021exploiting,pidhorskyi2020adversarial,richardson2021encoding,tov2021designing,wang2022high}, or use a hybrid approach leveraging both methods~\cite{zhu2016generative,zhu2020indomain}. For the encoder-based methods, pSp~\cite{richardson2021encoding} proposes a feature pyramid encoder into $\mathcal{W}+$ space. ReStyle~\cite{abdal2019image2stylegan} iteratively refines the predicted style latent through a few forward passes. However, these effective approaches are designed for 2D StyleGAN. Recently, IDE-3D~\cite{sun2022ide} propose an inversion approach for a 3D neural renderer with semantic masks yet is not able to generalize to several pre-trained style-based NeRFs. In this work, we would like to design a NeRF inversion model which is generalizable for most style-based NeRFs.

%% file: 3-method.tex
\section{The Proposed Approach}
\label{sec:method}

\subsection{Problem Formulation and Overview}

\textbf{Inversion of 2D generative model: }
In the encoder-based 2D GAN inversion, the goal is to train an encoder $E$ to generate the latent code $w$\footnote{latent $w \in \mathcal{W}$~\cite{shen2020interpreting} or the extended latent $w \in \mathcal{W}+$~\cite{Karras2019stylegan2}} for the given target image $x$ and minimize the distance between the input image and the generated image: 
\begin{equation}\label{eq:1}
\begin{split}
\min_{E} \mathcal{L}(x, G(w)), \text{ s.t. } w = E(x)
\end{split}
\end{equation}
where $G$ indicates the generator (\textit{i.e.,} StyleGAN). The objective can be $L_2$ distance, perceptual distance (LPIPS)~\cite{zhang2018unreasonable}, or a more sophisticated loss which consists of various reconstruction losses and regularization terms. We use an encoder to compute a latent code $w = E(x)$ to minimize the reconstruction loss. This allows fast inference without per-input optimization. 

\textbf{Inversion of style-based NeRF: }
For 3D style-based NeRF inversion, we not only need to reconstruct same-view images but also generate novel views of the same identity:
\begin{equation}\label{eq:1}
\begin{split}
\min_{E} \sum_{i=0}^{n} \mathcal{L}  {(x_i, G(w, p_i))}, \text{ s.t. } w = E(x_{0}), 
\end{split}
\end{equation}
where $x_i$, $i=0,1,..,n$ represent multi-view images that has the same identity as $x_{0}$ and $p_i$ are the corresponding camera poses. 
Minimizing the objective allows the model to learn the view-invariant latent code $\hat{w}$ since it maps multi-view images $x_i$ (controlled by the pose $p_i$) to the same latent code $w$ for each set of the training sample. During inference, a single-view image is mapped to the latent code $\hat{w}$ which can produce multi-view images of the same identity by changing the poses.

%
\begin{figure*}[t!]
  \centering
  \includegraphics[width=\linewidth]{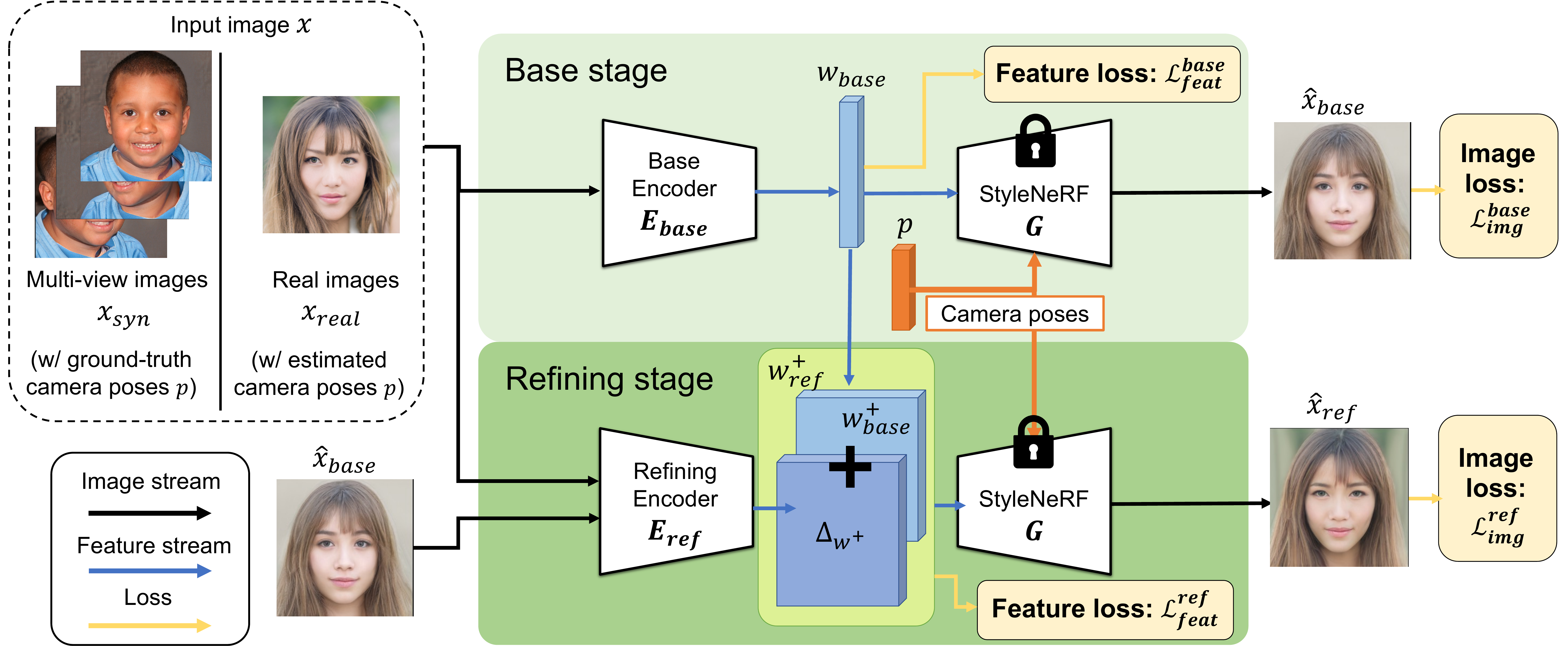}
  \vspace{-6mm}
  \caption{\textbf{Overview of our proposed 3D-aware Encoder for style-based NeRF: NeRF-3DE.} It consists of two stages: the base stage and the refining stage, and is trained with feature-level 3D-aware losses
  \protect\footnotemark and image-level reconstruction losses. More details can be referred to the section~\ref{sec:method}.}
  \label{fig:model}
  \vspace{-6mm}
\end{figure*}
\footnotetext{feature-level losses include triplet losses and L1 losses, which can only be applied to synthesized multi-view images in this work.}

\begin{figure}[t!]
  \centering
  \includegraphics[width=0.8\linewidth]{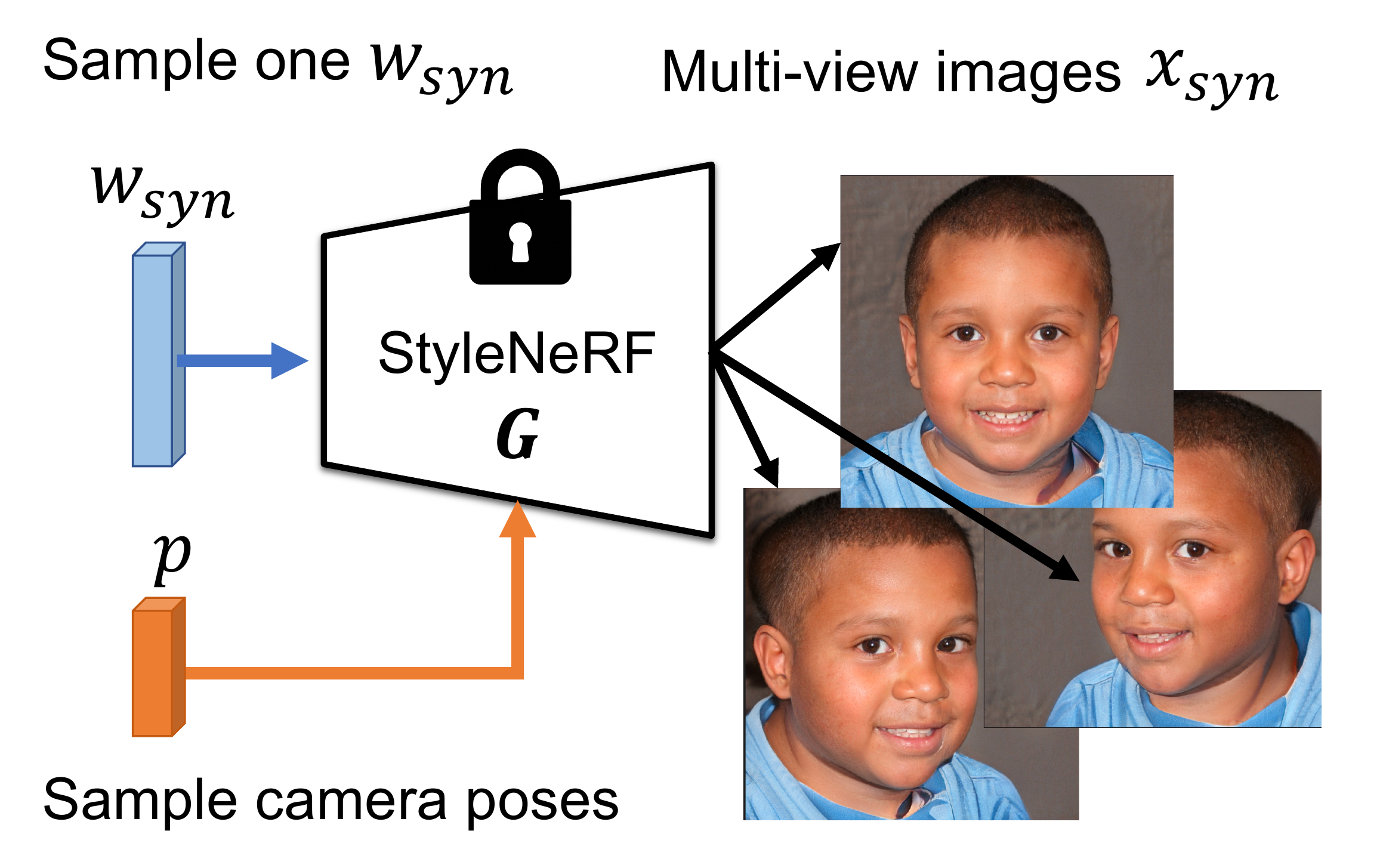}
  \vspace{-5mm}
  \caption{{The process of generating multi-view images by feeding the same latency code with different camera poses to StyleNeRF.}}
  \label{fig:syn}
  \vspace{-4mm}
\end{figure}


\textbf{Method overview}: In order to perform NeRF inversion for style-based NeRFs, we propose an encoder-based framework named NeRF-3DE, and the overview of the pipeline is presented in Figure~\ref{fig:model}.
The NeRF-3DE involves two stages: the base stage and the refining stage. 1) In base stage, the introduced base encoder $E_{base}$ takes an image $x$ as input and produces the style latent code $w_{base}$. 
In order to learn the view-invariant latent code, we leverage the multi-view images synthesized by the generator, shown in Figure~\ref{fig:syn}. This latent code $w_{base}$ is optimized to roughly reconstruct the 2D input image $\hat{x}_{base}$ and enable 3D-consistent novel-view rendering. 
2) To further minimize the identity gap between the output and the input images, a refining encoder $E_{ref}$ is introduced to refine the latent code $w_{base}$. It first takes the concatenation of the input image $x$ and the generated image $\hat{x}_{base}$ from the previous stage as input and learns a residue $\Delta_{w^+}$. Then we can obtain the output style latent code $\hat{w} = w_{ref}^+$ by adding the residue to $w^{+}_{base}$.

\subsection{Preliminary of Style-based NeRF (StyleNeRF)}
\paragraph{Style-based Neural Radiance Field.} Following StyleGANs~\cite{karras2019style,Karras2019stylegan2}, StyleNeRF~\cite{gu2021stylenerf} also introduce the mapping network $f$ which maps noise vectors from a spherical Gaussian space $\mathcal{Z}$ to the style space $\mathcal{W}$. $f$ consists of several MLP layers and the input style vector $w \in \mathcal{W}$ can be derived by $w = f(z), z \in \mathcal{Z}$. Following the neural rendering mechanism in NeRF~\cite{mildenhall2020nerf}, our model also takes the position $u\in \mathbb{R}^3$ and viewing direction $d \in \mathbb{S}^2$ as inputs, and predicts the density $\sigma(u)\in \mathbb{R}$ and view-dependent color $c(u,d) \in \mathbb{R}^3$.

In order to render the color and density for each coordinate in 3D space with high-frequency details, StyleNeRF also uses positional embedding with Fourier series: $\gamma(p) = (\sin(2^0\pi p), \cos(2^0\pi p), ... , \sin(2^{L-1}\pi p), \cos(2^{L-1}\pi p))$,
where the function $\gamma(.)$ is applied to each of the three coordinates in $u$ and to the three coordinates of the view direction $d$. Let us denote the rendering network by $\phi^n_{w}$ where $n$ indicates the number of MLP layers within and $w$ indicates the style feature. Each MLP weight matrix is modulated by the latent code $w$ independently. Both the density and the color can be rendered respectively with:
\begin{equation}\small
\begin{split}
\sigma_{w}(x) = h_{\sigma}(\phi^n_{w}(\gamma(u)))\\
c_{w}(u,d) = h_{c}(\phi^n_{w}(\gamma(u)),\gamma(d)),
\end{split}
\end{equation}
where $h_{\sigma}(\cdot)$ and $h_{c}(\cdot)$ are projection layers.

\paragraph{Volume Rendering with Radiance Fields.}
Once we have the color and density for each coordinate and view direction, we render the color $C(r)$ for each pixel along that camera ray $r(t)=o+td$ passing through the camera center $o$ with volume rendering~\cite{kajiya1984ray}:
\begin{equation}\small
\begin{split}
C_{w}(r) = \int_{t_n}^{t_f} T(t)\sigma_{w}(r(t))c_{w}(r(t),d)dt,\\
\text{where}\quad T(t) = \exp(-\int_{t_n}^t \sigma_{w}(r(s))ds).
\end{split}
\end{equation}
The function $T(t)$ denotes the accumulated transmittance along the ray from $t_n$ to $t$. In practice, the continuous integration is discretized by accumulating sampled points along the ray. More details can be obtained in NeRF~\cite{mildenhall2020nerf} and StyleNeRF~\cite{gu2021stylenerf}.

\subsection{Inversion of the view-invariant latent in W}
Unlike 2D GAN inversion which only generates the output image with the same camera pose as the input image, the inversion of 3D generative NeRF has to consider the optimization of unseen views for the input image. However, the latent code $w \in \mathcal{W}$ obtained by training on single-view images may not lead to high-quality novel-view images. See Figure \ref{fig:teasor} (c).

In order to learn the 3D-aware latent code $w$, we introduce a base encoder $E_{base}$ that is able to generate view-invariant latent code. In other words, for multi-view images of the same identity, we hope $E_{base}$ to map them to the same latent code: $w_{base}^{(i)} = E_{base}(x_j^{(i)})$, where $x_j^{(i)}$ denotes an image corresponding to camera pose $p_j$ of the identity-$i$.
To ensure this, we can use contrastive learning to train the encoder. 
%
Specifically, we perform contrastive learning with triplet loss $\mathcal{L}_{tri}$ on the feature vector $w$, which would maximize the inter-class discrepancy while minimizing intra-class distinctness. Specifically, for each input image $x$, we sample a positive image $x_\mathrm{pos}$ with the same identity label and a negative image $x_\mathrm{neg}$ with different identity labels to form a triplet tuple. Then, the following equations compute the distances between $x$ and $x_\mathrm{pos}$/$x_\mathrm{neg}$:

\vspace{-0.2cm}
\begin{equation}\small
  \begin{aligned}
  d_\mathrm{pos} = \|{w}_{base} - {w}_{{base}_\mathrm{pos}}\|_2, \quad d_\mathrm{neg} = \|{w}_{base} - {w}_{{base}_\mathrm{neg}}\|_2,
  \end{aligned}
  \label{eq:d-pos}
\end{equation}
where ${w}_{base}$, ${w}_{{base}_\mathrm{pos}}$, and ${w}_{{base}_\mathrm{neg}}$ represent the feature vectors of images $x$, $x_\mathrm{pos}$, and $x_\mathrm{neg}$, respectively. With the above definitions, we have the triplet loss $\mathcal{L}_{tri}$ defined as
\begin{equation}\small
  \begin{aligned}
  \mathcal{L}_{tri} (w_{base}) &~ 
  = \max(0, m + d_\mathrm{pos} - d_\mathrm{neg}),
  \end{aligned}
  \label{eq:tri}
\end{equation}
where $m > 0$ is the margin used to define the distance difference between the positive image pair $d_\mathrm{pos}$ and the negative image pair $d_\mathrm{neg}$. Contrastive learning requires multi-view images with the same identities. In reality, collecting such datasets is nontrivial, as it requires synchronized and calibrated camera arrays. To bypass this, we utilize images synthesized by the generator (\textit{i.e.,} StyleNeRF) itself. We can sample latent codes $w_{syn}$ from a StyleNeRF's $\mathcal{W}$ space, then sample different camera poses to generate multi-view images of the same identities as $x_{syn}$. 

Since we have the $w_{syn}$ latent code for $x_{syn}$, we can directly apply an $L_1$ loss between the predicted $w_{base}$ and the ``ground-truth'' $w_{syn}$.
The feature-level loss for synthesized images is summed up as:
\begin{equation}\small
  \begin{aligned}
  \mathcal{L}_{feat}^{base} &~ 
  = \mathcal{L}_{tri}(w_{base}) + \mathcal{L}_{1}(w_{base},w_{syn}),
  \end{aligned}
  \label{eq:feat}
\end{equation}

On the other hand, we are able to utilize both the real images $x_{real}$ and synthesized images $x_{syn}$ to train the base encoder with image-level loss. We construct the image-level loss using the pixel-wise $L_2$ loss and LPIPS loss~\cite{zhang2018unreasonable}. Following pSp~\cite{richardson2021encoding},  we also apply an identity (ID) similarity loss by employing a pre-trained facial recognition ResNet-IRSE50~\cite{deng2019arcface} to measure the facial identity:
\begin{equation}\small
  \begin{aligned}
  \mathcal{L}_{img}^{base} &~ 
  = \mathcal{L}_{2}(\hat{x}_{base},x)+\mathcal{L}_\mathrm{LPIPS}(\hat{x}_{base},x)
  +\mathcal{L}_{ID}(\hat{x}_{base},x),
  \end{aligned}
  \label{eq:syn_base}
\end{equation}
where $\hat{x}_{base} = G(E_{base}(x),p)$ and $p$ indicates the corresponding camera pose. We use the ground truth camera poses for synthesized images and camera poses predicted by the off-the-shelf predictor~\cite{ruiz2018fine} for real images. 
The image-level losses can be summed up for both synthesized images and real images. 
With the image-level loss $\mathcal{L}_{img}^{base}$, the base encoder is able to learn to reconstruct the images by back-propagating through the generator.

\subsection{Refinement of the latent in W+}


While $w$ latent code is learned to preserve the 3D structure with our base encoder $E_{base}$, it leads to poor identity preservation. Thus, we introduce a refining encoder to refine the latent code $w_{base}$.

Following previous works for learning the $\mathcal{W}+$ latent instead of $\mathcal{W}$, we first duplicate the base latent $w_{base} \in \mathbb{R}^{d}$ to $w_{base}^+ \in \mathbb{R}^{n\times d}$ and learn the refined $w_{ref}^+$ by adding the learned residue $\Delta_{w^+}$:
\begin{equation}\small
  \begin{aligned}
   w_{ref}^+ = w_{base}^+ + \Delta_{w^+}\quad \text{and}\quad \Delta_{w^+} = E_{ref}(x,\hat{x}_{base}),
  \end{aligned}
  \label{eq:resi_latent}
\end{equation}
where $w_{ref}^+$ is in the $\mathcal{W}+$ latent space and is capable of better reconstructing the input image using the generator $G$. In order to learn the residue $\Delta_{w^+}$, we introduce the refining encoder $E_{ref}$ which takes the input image x and the previously generated image $\hat{x}_{base}$ as inputs and produces $\Delta_{w^+}$.
The design of the refining stage is similar to the ReStyle~\cite{alaluf2021restyle} originally proposed for StyleGAN. The difference lies in that Restyle~\cite{alaluf2021restyle} uses the randomly averaged $w$ latent code and the corresponding synthesized image as inputs while our $E_{ref}$ uses the outputs  (i,e, $w_{base}$ and $x_{base}$) of the base encoder. Same as the base stage, we employ the same image-level losses to train the refining encoder:
\begin{equation}\small
  \begin{aligned}
  \mathcal{L}_{img}^{ref} &~ 
  = \mathcal{L}_{2}(\hat{x}_{ref},x)+\mathcal{L}_\mathrm{LPIPS}(\hat{x}_{ref},x)
  +\mathcal{L}_{ID}(\hat{x}_{ref},x),
  \end{aligned}
  \label{eq:syn_res}
\end{equation}
where $\hat{x}_{ref} = G(E_{ref}(x,\hat{x}_{base}),p)$ and $p$ indicate the corresponding camera poses similar to base stage. $x$ can either be a real image or a synthetic image. Since we also have ground truth $w_{syn}^+$ from synthesized images, we can also train the refining encoder using the feature-level $L_1$ loss:
\begin{equation}\small
  \begin{aligned}
  \mathcal{L}_{feat}^{ref} &~ 
  = \mathcal{L}_{1}(w_{ref}^+,w_{syn}^+),
  \end{aligned}
  \label{eq:feat}
\end{equation}
Note that we do not use the triplet loss on the refining latent code $w_{ref}^+$ since the large dimension size of the $\mathcal{W}+$ space ($\mathbb{R}^{n\times d}$) makes the contrastive learning prone to overfitting. 

Similar to Restyle~\cite{alaluf2021restyle}, our refining encoder can also perform multiple iterative refinements using Equation~\ref{eq:resi_latent}.

The total loss $\mathcal{L}$ for training our proposed NeRF-3DE is summarized as follows:
\begin{equation}\small
  \begin{split}
  \mathcal{L}_{total} & = \lambda_{feat}^{base}\cdot\mathcal{L}_{feat}^{base} + \lambda_{img}^{base}\cdot\mathcal{L}_{img}^{base}\\ & + \lambda_{feat}^{ref}\cdot\mathcal{L}_{feat}^{ref} + \lambda_{img}^{ref}\cdot\mathcal{L}_{img}^{ref},
  \end{split}
  \label{eq:fullobj}
\end{equation}
where $\lambda_{feat}^{base}$, $\lambda_{img}^{base}$, $\lambda_{feat}^{ref}$ and $\lambda_{img}^{ref}$ are the hyper-parameters used to control the weighting of the corresponding losses.

%% file: 4-experiment.tex
\section{Experiment}
\label{sec:EXP}

\begin{figure*}[t!]
  \centering
  \includegraphics[width=0.75\linewidth]{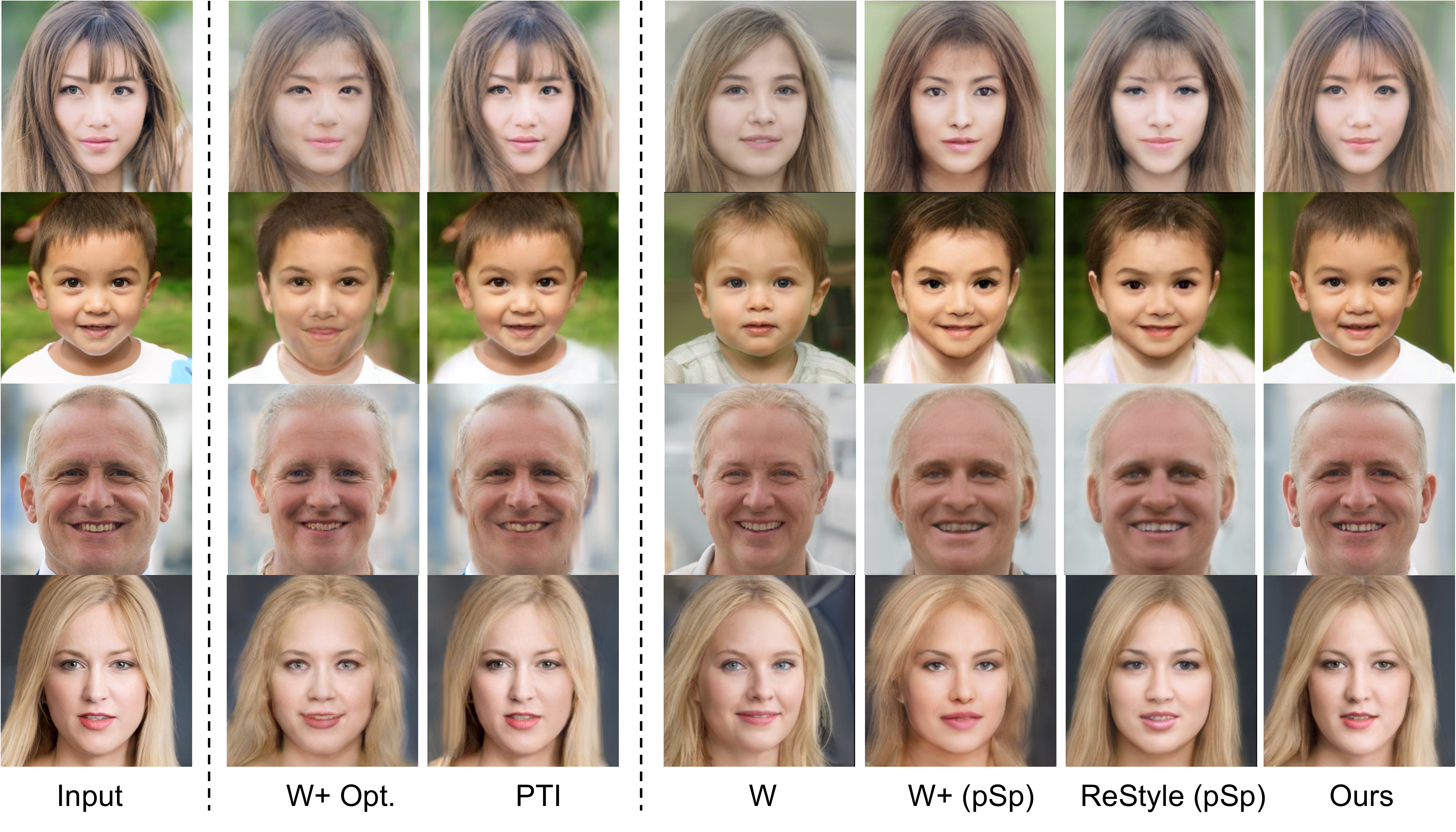}
  \vspace{-5mm}
  \caption{\textbf{Qualitative comparisons on image reconstruction.} All of the output images are rendered using the same camera pose as the input image from the StyleGAN2-Fake dataset.}
  \label{fig:qual}
\end{figure*}
\input{exp/quant_metric}
\begin{figure*}[t!]
  \centering
  \includegraphics[width=0.75\linewidth]{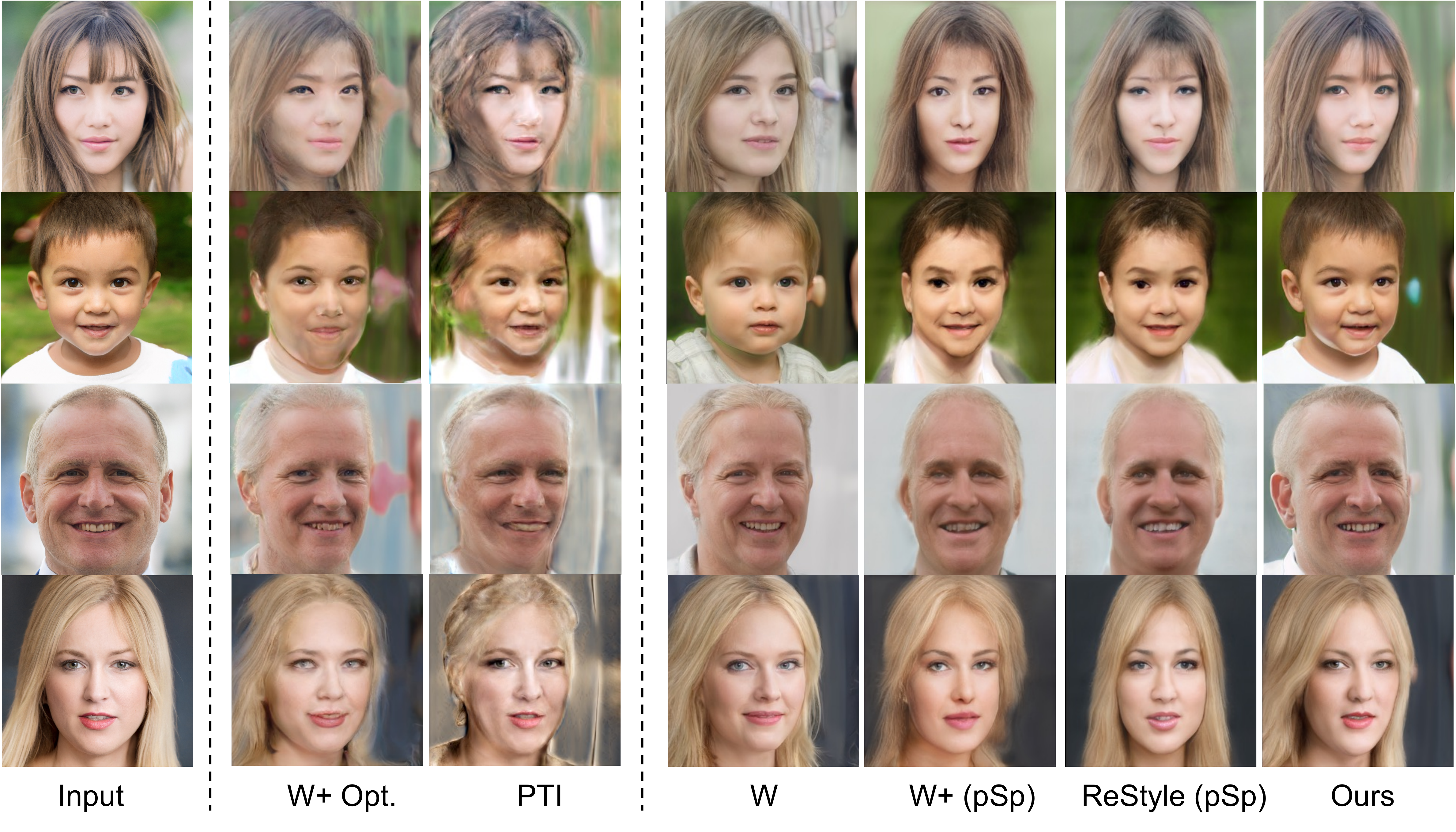}
  \vspace{-5mm}
  \caption{\textbf{Qualitative comparisons on novel-view rendering.} All of the output images are rendered using face yaw angle $-35^\circ$ degrees. The images are from the StyleGAN2-Fake dataset.}
  \label{fig:novel}
  \vspace{-3mm}
\end{figure*}
\input{exp/quant_novel}
\subsection{Experimental Settings}

\paragraph{Datasets.}
By default, all our experiments are conducted on human faces and using StyleNeRF~\cite{gu2021stylenerf} as the pretrained generator for GAN inversion. We train the encoder for StyleNeRF using real images in \textbf{FFHQ}~\cite{karras2019style} (i.e., the same dataset used for StyleNeRF training) and multi-view synthesized images from StyleNeRF itself.
We use the \textbf{CelebA-HQ} test set~\cite{karras2017progressive,liu2015deep} for quantitative evaluations.
For qualitative evaluation, we visualize the results using images from the \textbf{StyleGAN2-Fake} dataset as inputs, which has $263$ curated images generated and released by StyleGAN2~\cite{Karras2019stylegan2} to avoid showing real faces. The camera poses for all real input are derived using the off-the-shelf pose estimator: HopeNet~\cite{ruiz2018fine} for a fair comparison with previous works.
For synthesized images, we use their ground truth camera poses for both training and inference. 
Moreover, we will present results on animal faces on AFHQ dataset\cite{choi2020starganv2} and the results based on EG3D (trained with FFHQ or AFHQ-cat and its self-generated images) in the supplementary material.

\paragraph{Baselines.}
Since our NeRF-3DE is the first 3D-aware encoder for style-based NeRFs, we compare it with several baselines. The first set of baselines is directly built from current state-of-the-art styleGAN inverters, including pSp~\cite{richardson2021encoding} and ReStyle~\cite{alaluf2021restyle}. We also build a baseline encoder for $\mathcal{W}$ inversion.
For a fair comparison, all of the encoder-based competitors are trained on the same dataset, \textit{i.e.}, using both real and synthesized images. 
%
To compare with online optimization methods, we compare our model with latent vector optimization in $\mathcal{W}+$~\cite{Karras2019stylegan2}  and PTI~\cite{roich2021pivotal}.


\paragraph{Evaluation settings.}
We conduct the experiments in two settings: 1) Same-view image reconstruction and 2)  Novel-view image rendering.
%
%
\begin{figure*}[t!]
  \centering
  \includegraphics[width=0.75\linewidth]{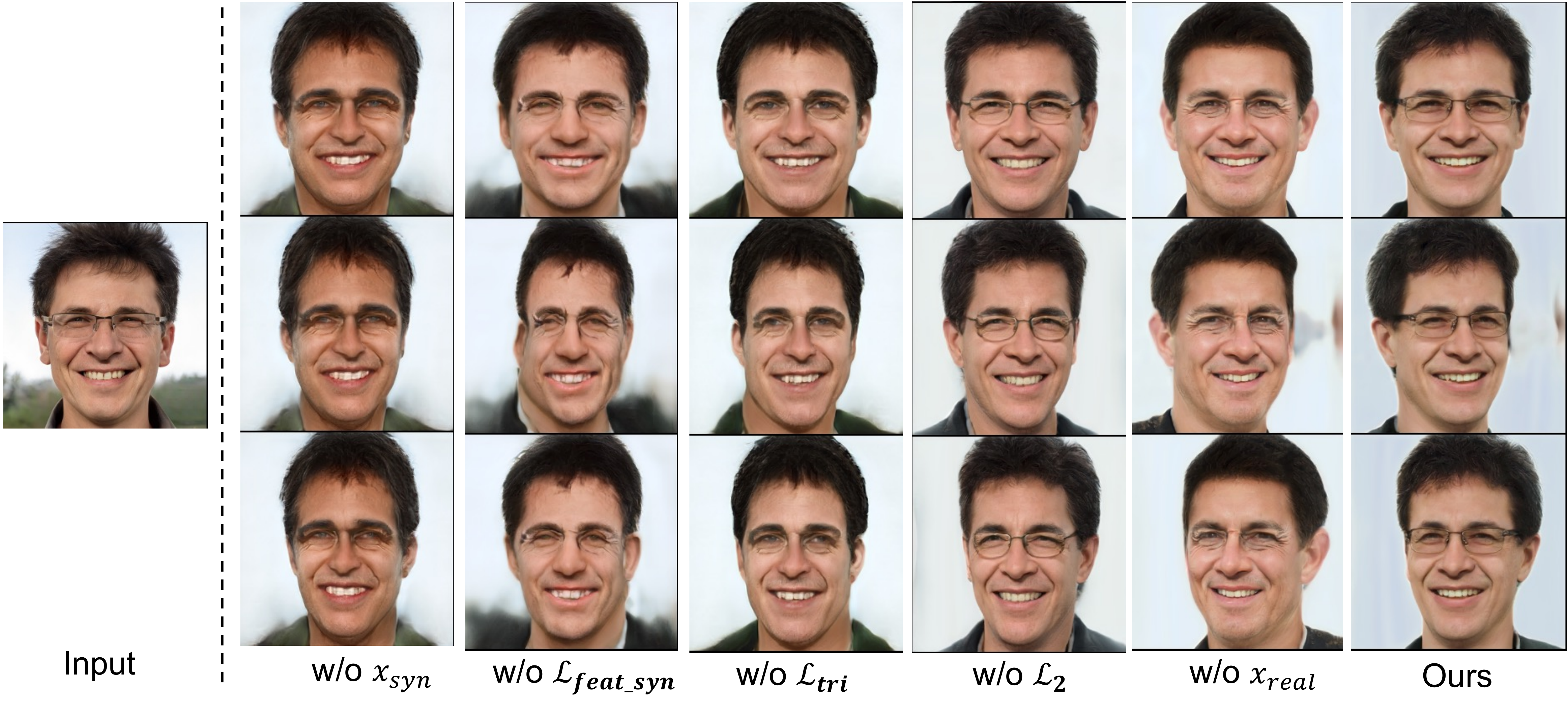}
  \vspace{-5mm}
  \caption{\textbf{Ablation studies on the training images and the feature losses.} The testing images are from StyleGAN-Fake.}
  \label{fig:abl}
  \vspace{-4mm}
\end{figure*}
For the first setting, we visually compare the input image and the corresponding output image generated from the latent code and the camera pose of the input image. We also quantitatively evaluate the distance between the input and output images using the metrics: $L_2$, LPIPS~\cite{zhang2018unreasonable}, 
and identity (ID)~\cite{huang2020curricularface}. For the second setting, we qualitatively and quantitatively compare the input image and the novel views
(e.g., $-35^{\circ}$ yaw angle)
image generated from its latent code. Since we do not have the ground truth to measure the distance, we only quantitatively evaluate the identity (ID) distance~\cite{huang2020curricularface} with input from different views. We would like to note that, the original generator (StyleNeRF~\cite{gu2021stylenerf}) is trained using the head yaw angle ranging between $-17^{\circ}$ to $17^{\circ}$ degrees. 
Based on our observation, the pretrained StyleNeRF itself can only generate images at most twice the yaw range (i.e., $-35^{\circ}\sim+35^{\circ}$) before breaking the 3D structure. Thus, we set the rendering yaw range of our NeRF-3DE to $-35^{\circ}\sim+35^{\circ}$ and the default roll angle as $0^{\circ}$.


\subsection{Results of image reconstruction}

In this section, we compare our proposed model with three encoder-based models and two optimization approaches quantitatively (in Table~\ref{table:abl_rec}) and qualitatively (in Figure~\ref{fig:qual}). 
As listed in Table~\ref{table:abl_rec}, among all encoder-based methods, our proposed method achieves the best results across all four metrics. For example, it outperforms the previous effective StyleGAN encoder, i.e., ReStyle~\cite{or2022stylesdf}, with a large gap
and outperforms the other two baselines even more, which demonstrates that it is not optimal to directly apply existing encoders to invert style-based NeRFs. Visually, our model also greatly outperforms all encoder-based methods as shown in Figure~\ref{fig:qual}. For example, although all models can generate realistic faces due to the pretrained StyleNeRF, our proposed model can better reconstruct the input image which is consistent with our quantitative results. To dig deeper, we observe that although encoding in $\mathcal{W}$ achieves the worst image reconstruction results, it has much better 3D preservation than encoding the latent in $\mathcal{W+}$, which we will discuss in the next section (see column 5 \& 6 vs column 4 in Figure~\ref{fig:qual} and Figure~\ref{fig:novel}. This is also what motivates us to build our base encoder to learn latent code in $\mathcal{W}$ space rather than $\mathcal{W+}$ space.

Besides encoder-based baselines, we also compare our proposed model with the 2D online optimization methods (see rows 2 \& 3 in Table~\ref{table:abl_rec}). The online optimization methods are much slower than encoder-based methods, and in return, their performance for image reconstruction is known to be the upper bound for that of encoder-based methods~\cite{alaluf2021restyle,richardson2021encoding}.
Our model outperforms the online method $\mathcal{W+}$ Opt. (see Table~\ref{table:abl_rec} row 3) and $\mathcal{W+}$ Opt. does not perfectly reconstruct the input image although it works effectively in 2D StyleGAN. PTI still has the best performance for image reconstruction with 3D style-based NeRF quantitatively and qualitatively. However, we greatly reduced the gap between encoder-based methods and the online optimization method for same-view image reconstruction.

\begin{figure}[t!]
  \centering
  \includegraphics[width=\linewidth]{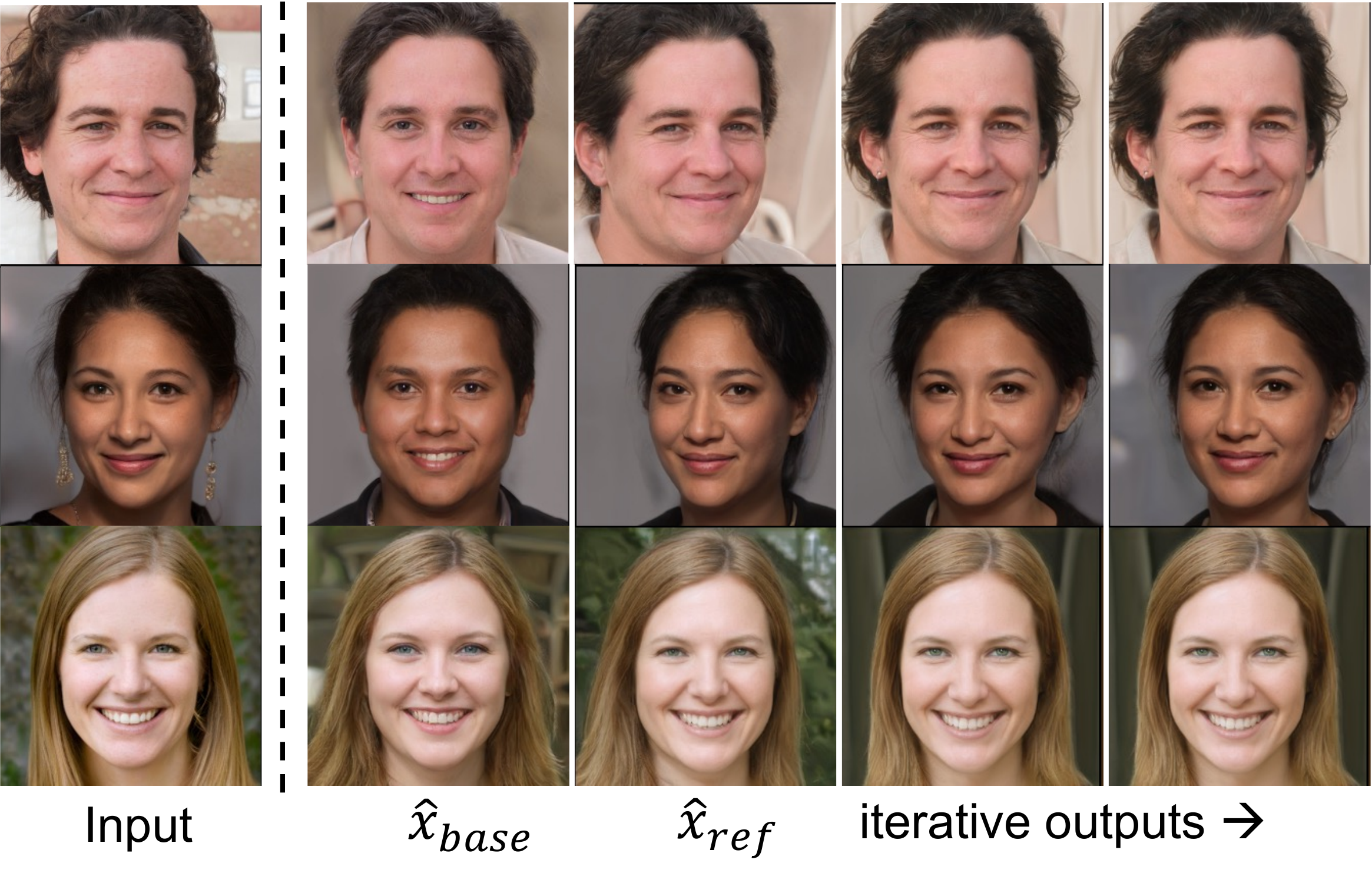}
  \vspace{-8mm}
  \caption{\textbf{Ablation studies on the outputs of each stage.} The testing images are from StyleGAN-Fake.}
  \label{fig:iter}
  \vspace{-4mm}
\end{figure}
\subsection{Results of novel-view rendering}

We present our qualitative result of novel viewing rendering of yaw angle $-35^\circ$ degree in Figure~\ref{fig:novel}, and also compare it with the same encoder-based models and optimization approaches.
First, comparing with encoder-based models for $\mathcal{W}$, $\mathcal{W+}$ using pSp~\cite{richardson2021encoding} and using ReStyle~\cite{alaluf2021restyle}, we found that our proposed NeRF-3DE not only effectively preserves the fine details and identity from the input image, but also maintains a reasonable 3D shape. Second, we observe that while encoding the latent in $\mathcal{W+}$ is more effective than encoding in $\mathcal{W}$ in image reconstruction, it generates inaccurate face angle or loses 3D preservation for novel views (see column 5 \& 6). On the other hand, encoding latent in $\mathcal{W}$, though with much worse identity preservation, has better 3D preservation and correct view-angle. Third, the optimization method PTI~\cite{roich2021pivotal} though has near-perfect image reconstruction of the same view, it breaks 3D structure when rendering novel views (See column 3 of Figure~\ref{fig:qual} and Figure~\ref{fig:novel}). Consequently, compared with all baselines including online optimization and encoder-based models, our proposed NeRF-3DE achieves superior results in novel-view rendering.

We also benchmark the quantitative results using identity metrics for the selected four yaw head angles. As shown in Table~\ref{table:abl_nov}, the score usually decreases when the yaw head pose is more extreme. Our proposed model achieves the highest ID score among all competitors.
The optimization method PTI~\cite{roich2021pivotal}, though has achieved the highest scores in all of the evaluation metrics in Table~\ref{table:abl_rec}, also exhibits an inferior ID score to our encoder-based model.


\subsection{Ablation studies}


To further analyze the effectiveness of essential components of the proposed method, we conduct the experiments with one of them excluded 
and present the qualitative result in Figure~\ref{fig:abl}. When the synthesized images $x_{syn}$ are excluded (note that the feature losses $\mathcal{L}_{feat}$ will also be excluded without $x_{syn}$), the style latent generated by our proposed encoders could not preserve the reasonable face structure (see column 2). When using synthesized images $x_{syn}$ and only excluding $\mathcal{L}_{feat}$, the generated multi-view images still have artifacts and distortions in the face (see column 3). In addition, we found that $\mathcal{L}_{tri}$ in $\mathcal{L}_{feat}$ serves as a more important role for the 3D and identity preservation (see column 4 vs column 5). Moreover, if the encoders are trained with synthesized images $x_{syn}$ only (\textit{i.e.}, w/o $x_{real}$), the style latent code is able to preserve view consistency yet still has the loss of identity preservation compared with the full model (see column 6 vs column 7). These studies demonstrate that both real images and synthesized images with feature-level losses are significant to our model.

To further analyze the importance of both our base encoder and the refining encoder, we also visualize the output of these two encoders. The visualization is presented in Figure~\ref{fig:iter}. The output of $\hat{x}_{base}$ can be seen as using the encoding into $\mathcal{W}$ in the fourth column of Figure~\ref{fig:novel} plus the feature-level loss. Though the generated $\hat{x}_{base}$ has a gap from the input image, it preserves 3D view consistency. Then we can produce the latent code for generating $\hat{x}_{ref}$ on top of $\hat{x}_{base}$ with more fine details. We also demonstrate that our restyle steam can also be done in several iterations yet does not improve the latent code as much as Restyle~\cite{alaluf2021restyle} presented in 2D StyleGAN.


\subsection{Extension to support online optimization}\label{app:online}
\begin{figure}[t!]
  \centering
  \includegraphics[width=0.95\linewidth]{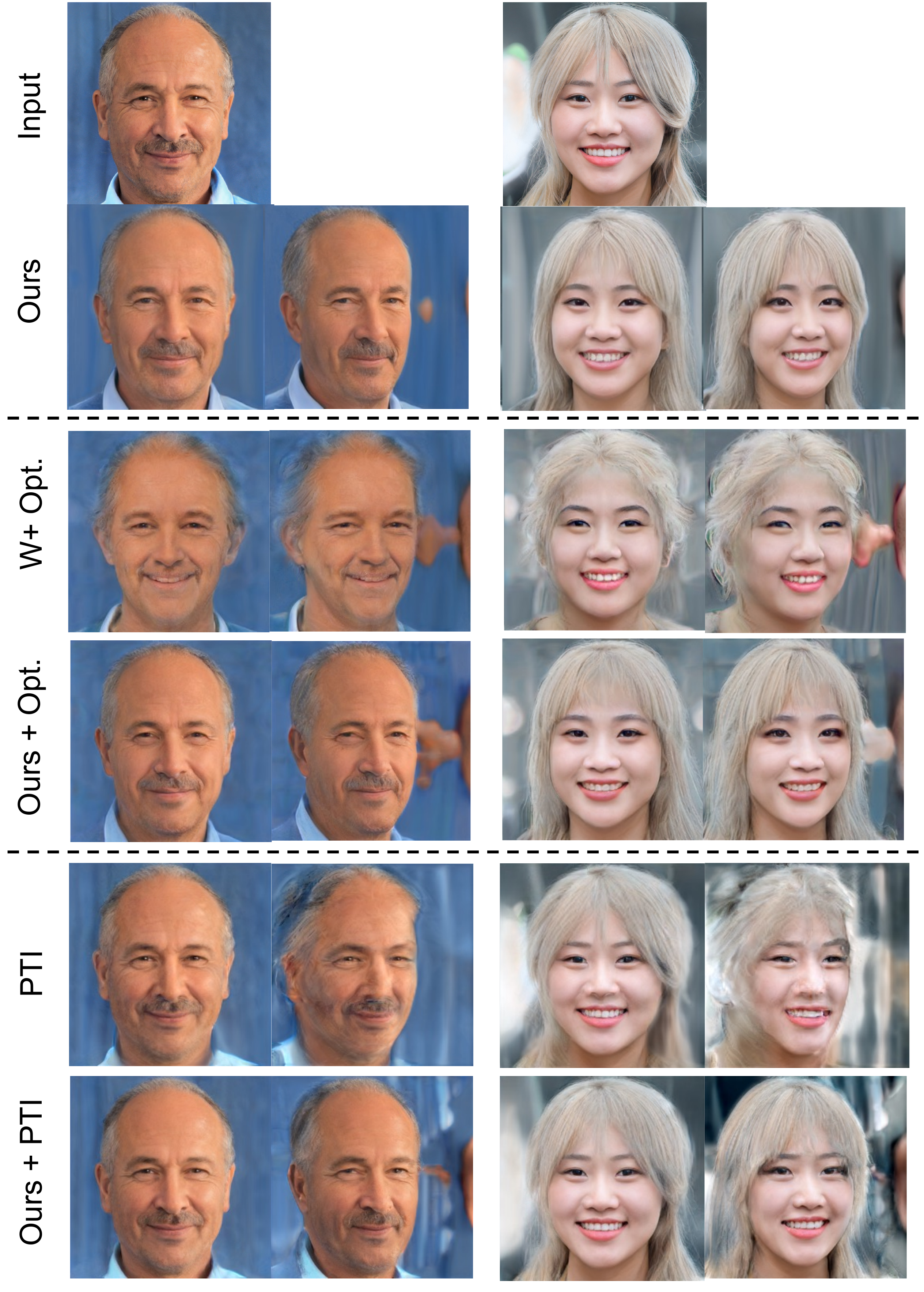}
  \vspace{-3mm}
  \caption{\textbf{The effectiveness of our model for online optimization.} We utilize the starting latent produced by our model for Opt. (optimization to $\mathcal{W}+$~\cite{Karras2019stylegan2}) and PTI~\cite{roich2021pivotal}.}
  \label{fig:online}
\end{figure}

In this section, we would like to analyze the effectiveness and the possibility of our encoders for supporting online optimization. We conduct the experiments of utilizing our model for producing initial style latent code in both latent vector optimization to $\mathcal{W}+$~\cite{Karras2019stylegan2} and PTI~\cite{roich2021pivotal}. The results and comparison are presented in Figure~\ref{fig:online}. We can observe that, for both of the optimization approaches, our encoders improve the identity and 3D preservation for both image reconstruction and novel-view rendering ($-35^\circ$ in the examples).

%% file: exp/quant_metric.tex
  
  

\begin{table}[t!]
  
  \centering
  
  \resizebox{\linewidth}{!}
  {
  \begin{tabular}{cl|ccccc}
  \toprule
  & {Method} & $\downarrow$ $L_2$ & $\downarrow$ LPIPS & $\uparrow$ ID & Time (s) $\downarrow$\\
  \midrule
  \multirow{2}{*}{Online-based} & PTI~\cite{roich2021pivotal} & 0.03 & 0.09 & 0.86 & 194.203 \\
  & $\mathcal{W}+$ Opt.~\cite{Karras2019stylegan2} & 0.08 & 0.28 & 0.65 &  66.153\\
  \midrule
  \multirow{4}{*}{Encoder-based} & $\mathcal{W}$ & 0.12 & 0.31 & 0.60 &  0.105\\
  & $\mathcal{W}+$ (pSp)~\cite{richardson2021encoding} & 0.14 & 0.29 & 0.63 &  0.132\\
  & Restyle (pSp)~\cite{alaluf2021restyle} & 0.09 & 0.27 & 0.68 &  0.454\\
  & NeRF-3DE (Ours) & \textbf{0.05} & \textbf{0.21} & \textbf{0.72}  & 0.315\\
  %
  %
  \bottomrule
  \end{tabular}
  }
  \vspace{-2mm}
  \caption{\textbf{Quantitative comparison on image reconstruction} with online-based (i.e., upper bound for reconstruction) and encoder-based methods on the CelebA-HQ test dataset.}\label{table:abl_rec}
\end{table}

%% file: exp/quant_novel.tex

\begin{table}[t!]
  
  \centering
  
  \resizebox{\linewidth}{!}
  {
  \begin{tabular}{cl|cccc|c}
  \toprule
  & \multirow{3}{*}{Method} & \multicolumn{5}{c}{$\uparrow$ ID} \\
  \cmidrule{3-7} 
  & & \multicolumn{4}{c|}{Yaw angle} &\multirow{2}{*}{Avg.} \\
  %
  \cmidrule{3-6} 
  & & $-35^\circ$ & $-17^\circ$ & $17^\circ$ & $35^\circ$ \\
  %
  %
  \midrule
  \multirow{2}{*}{Online-based} & PTI~\cite{roich2021pivotal} & 0.41 & 0.46 & 0.44 & 0.40 & 0.43\\
  & $\mathcal{W}+$ Opt.~\cite{Karras2019stylegan2} & 0.23 & 0.28 & 0.27 & 0.22 & 0.25\\
  \midrule
  \multirow{4}{*}{Encoder-based} & $\mathcal{W}$ & 0.17 & 0.19 & 0.19 & 0.15 & 0.18\\
  & $\mathcal{W}+$ (pSp)~\cite{richardson2021encoding} & 0.21 & 0.27 & 0.31 & 0.24 & 0.22\\
  & Restyle (pSp)~\cite{alaluf2021restyle} & 0.20 & 0.35 & 0.32 & 0.21 & 0.27\\
  & NeRF-3DE (Ours) & \textbf{0.49} & \textbf{0.53} & \textbf{0.53} & \textbf{0.50} & \textbf{0.51}\\
  \bottomrule
  \end{tabular}
  }
  \vspace{-2mm}
  \caption{\textbf{Quantitative results on novel-view rendering, and comparison with online-based and encoder-based methods.} The results are measured on the CelebHQ test dataset.}\label{table:abl_nov}
  \vspace{-4.5mm}
\end{table}

%% file: 5-conclusion.tex
\section{Conclusion}

We have unveiled the challenges of NeRF inversion for style-based NeRF and the limitations of the current encoder-based models through experiments. To tackle the issue, we propose an encoder-based framework named NeRF-3DE, which consists of a base encoder and a residual encoder, to perform NeRF inversion for the 3D generative radiance field. Compared with the current existing encoder-based methods for GAN inversion, our proposed model achieves more effective NeRF inversion for 3D generative NeRF and has satisfactory image quality for rendering novel views. We also demonstrate that the style latent code generated by our proposed model is able to serve as a good initial point for online optimization.

%% file: appendix.tex
\section{More details of datasets and settings}

\paragraph{FFHQ} The FFHQ~\cite{karras2019style} dataset contains 70,000 face images. It is only used for training the initial checkpoint for the generator ($\mathbf{G}$) and the encoders in our framework.

\paragraph{CelebA-HQ} CelebA-HQ~\cite{karras2017progressive,liu2015deep} contains 24,183 training face images and 2,824 testing images. For a fair comparison with previous inversion methods, we only use the test split 2,824 images for testing. In this paper, since all of the testing images from this dataset are from the real human face, we did not present the qualitative visualizations for privacy protection. We only present the quantitative comparisons in the paper.

\paragraph{StyleGAN2-Fake} In order to present the rendering results qualitatively without using real faces, we use the fake yet very realistic faces released by \cite{Karras2019stylegan2}. This dataset contains 263 images of resolution 1,024×1,024 of very realistic human faces generated by StyleGAN2~\cite{Karras2019stylegan2}. We present the testing results qualitatively using these images.

\paragraph{AFHQ} Besides the experiments of inversion on human faces, we also conduct the experiments on animal faces using AFHQ~\cite{choi2020starganv2} and present the results later in the appendix. This dataset contains 15,000 high-quality images at 512×512 resolution and includes three categories of animals which are cats, dogs, and wildlife. Each category has about 5000 images. For each category, the dataset split around 500 images as a test set and provide all remaining images as a training set.

\section{Implementation Details}

All training and testing images are resized to size $256 \times 256 \times 3$, denoting width, height, and channel respectively. The experimental style-based NeRF generator ($G$) employs the checkpoint of StyleNeRF~\cite{gu2021stylenerf} with dimension $256$. The base encoder $E_{base}$ employs a series of residual blocks and 1 linear projection layer. The residual encoder $E_{res}$ employs the architecture from pSp~\cite{richardson2021encoding} and we set the number of residual iterations as 3 in the experiments. We set the dimension of the latent code $w$ as 512 which is the same as the generator and the number of latent code of $w+$ as $17$ following the checkpoint from StyleNeRF~\cite{gu2021stylenerf}.
For the hyperparameter for all of the loss functions, all of losses are equally weighted ($\lambda_{feat}^{base}=1.0$, $\lambda_{img}^{base}=1.0$, $\lambda_{feat}^{res}=1.0$ and $\lambda_{img}^{res}=1.0$) for all the experiments. The batch size is set as $32$ where $16$ is for synthesized images and $16$ is for real images. In the $16$ synthesized images in each batch, we sample $4$ identity latent $w_{syn}$ from StyleNeRF~\cite{gu2021stylenerf} and each $w_{syn}$ samples $4$ camera poses (randomly and uniformly sample yaw angle in the range $-35^{\circ}$ to $+35^{\circ}$ and roll angle as $0$ for simplicity), which can be formulated into these $16$ synthesized images.  We optimize the network using Adam optimizer with the learning rate set as $0.0001$.
Each experiment is conducted on 1 Nvidia GPU A100 (80G) with a batch size of 32 and implemented in PyTorch. We now present more details about the model architecture below:

\paragraph{Generator (StyleNeRF)}
\begin{figure}[t]
  \centering
  \includegraphics[width=\linewidth]{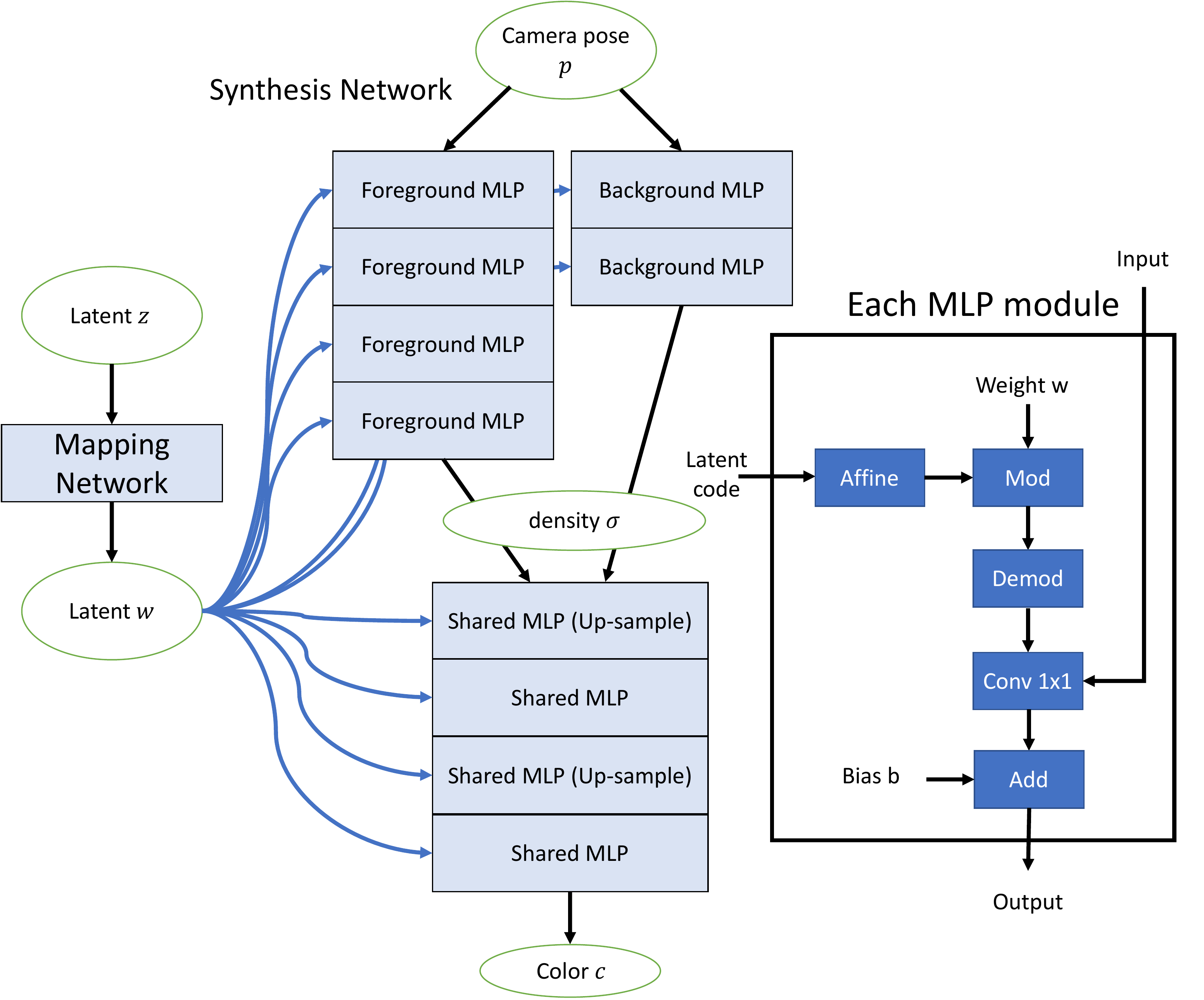}
  \caption{Brief overview of the architecture of StyleNeRF~\cite{gu2021stylenerf}.}
  \label{fig:stylenerf}
\end{figure}
StyleNeRF~\cite{gu2021stylenerf} has a mapping network and a synthesis network as StyleGAN~\cite{karras2019style} does. The overview of the network is roughly presented in Figure~\ref{fig:stylenerf}. For the mapping network, latent codes are sampled from standard Gaussian distribution and processed by a number of fully connected layers. The synthesis network employs
NeRF++ which consists of a unit sphere for foreground NeRF  and a background NeRF using inverted sphere parameterization. Two MLPs that represent foreground and background are used to predict the density. The color prediction is performed using another shared MLP. Each style-conditioned MLP block consists of an affine transformation layer and a 1×1 convolution layer. The convolution weights are modulated with the affine-transformed styles and then demodulated for computation. Leaky-ReLU is used as non-linear activation. We directly utilize the checkpoint provided by StyleNeRF~\cite{gu2021stylenerf} without further change on the network and the pre-trained weights. More details can be found at \cite{gu2021stylenerf}.

\begin{figure}[t]
  \centering
  \includegraphics[width=\linewidth]{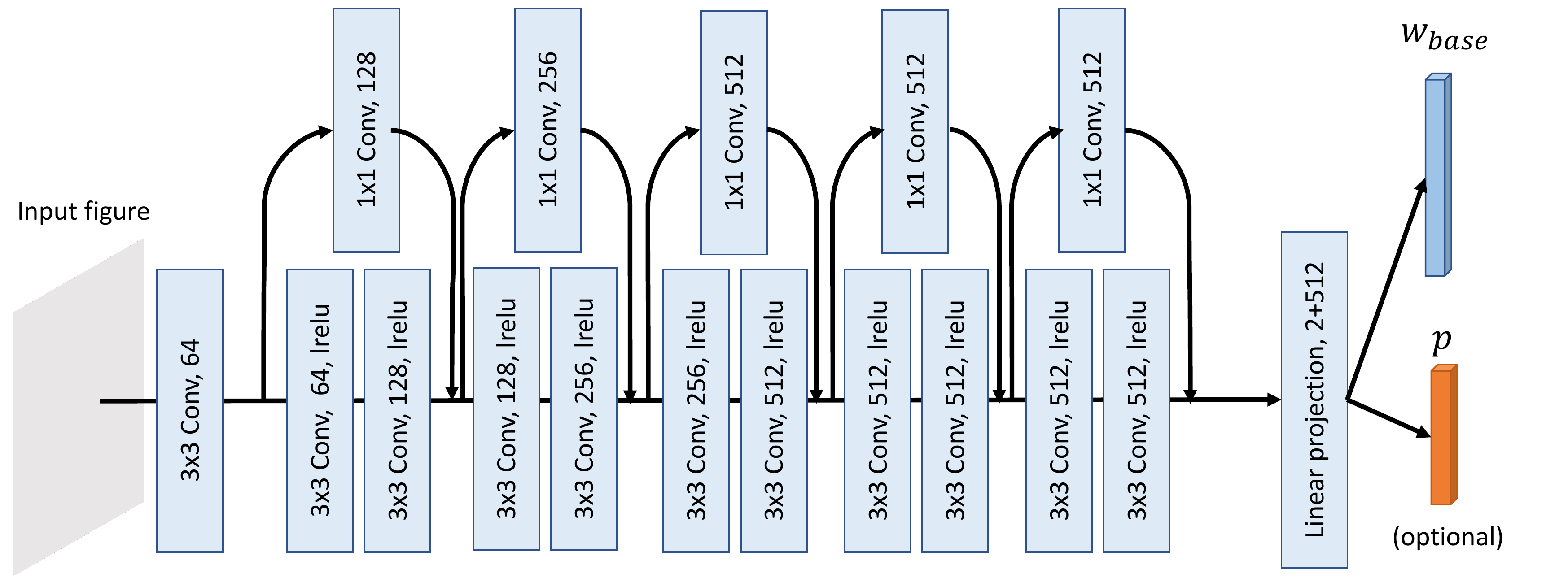}
  \caption{Brief overview of the architecture of base encoder.}
  \label{fig:encoder}
\end{figure}

\paragraph{Base Encoder}
As mentioned earlier, the base encoder $E_{base}$ contains 6 residual blocks and 1 linear projection layer. The output of the encoder will be a vector of 512-dimension $w$ latent code. The network is roughly presented in Figure~\ref{fig:encoder}. Since not all of the testing data in the real world has ground truth pose from the off-the-shelf model, our base encoder can also predict the yaw and roll angles from the input image while training with the ground-truth pose outputs from the synthesized images. The output dimension will be 514 (512 plus 2) if the additional task for pose prediction is added.

\begin{figure}[t]
  \centering
  \includegraphics[width=0.8\linewidth]{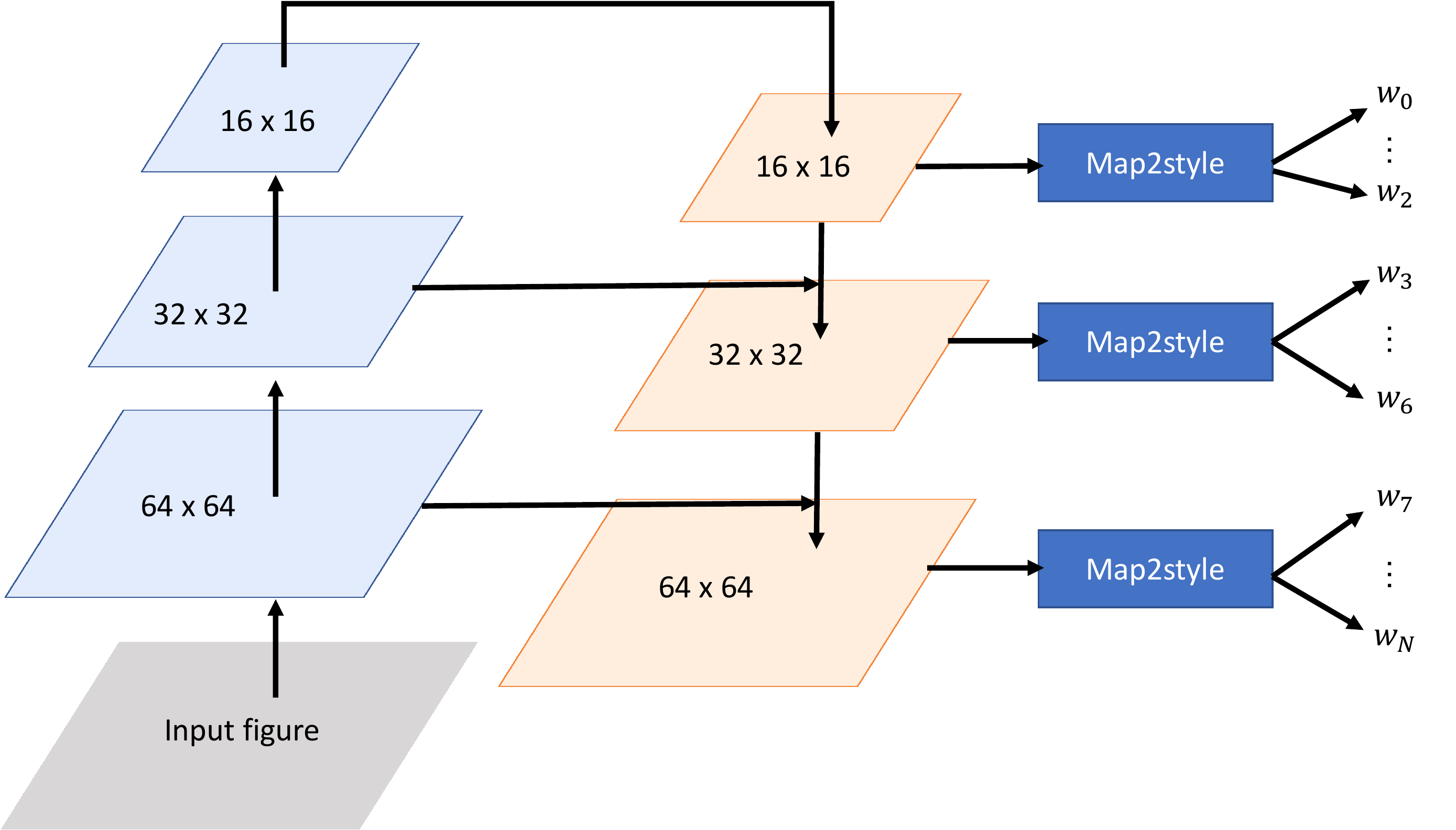}
  \caption{Brief overview of the architecture of residual encoder using pSp~\cite{richardson2021encoding}.}
  \label{fig:pSp}
\end{figure}

\paragraph{Residual Encoder}
The overview of the network is roughly presented in Figure~\ref{fig:pSp}.  The encoder derives the style input latent codes from three intermediate feature maps of spatial resolutions 16 × 16 (for input index 0 to 2), 32 × 32 (for input index 3 to 6), and 64 × 64 (for index 7 to last one). Each style vector is obtained from the corresponding feature map using a Map2style block, which is a convolutional network containing a series of 2-strided convolutions with LeakyReLU activations. More details can be found at \cite{richardson2021encoding}.

\section{More ablation studies and results}
\label{app:abl}

\begin{figure*}[t]
  \centering
  \includegraphics[width=0.9\linewidth]{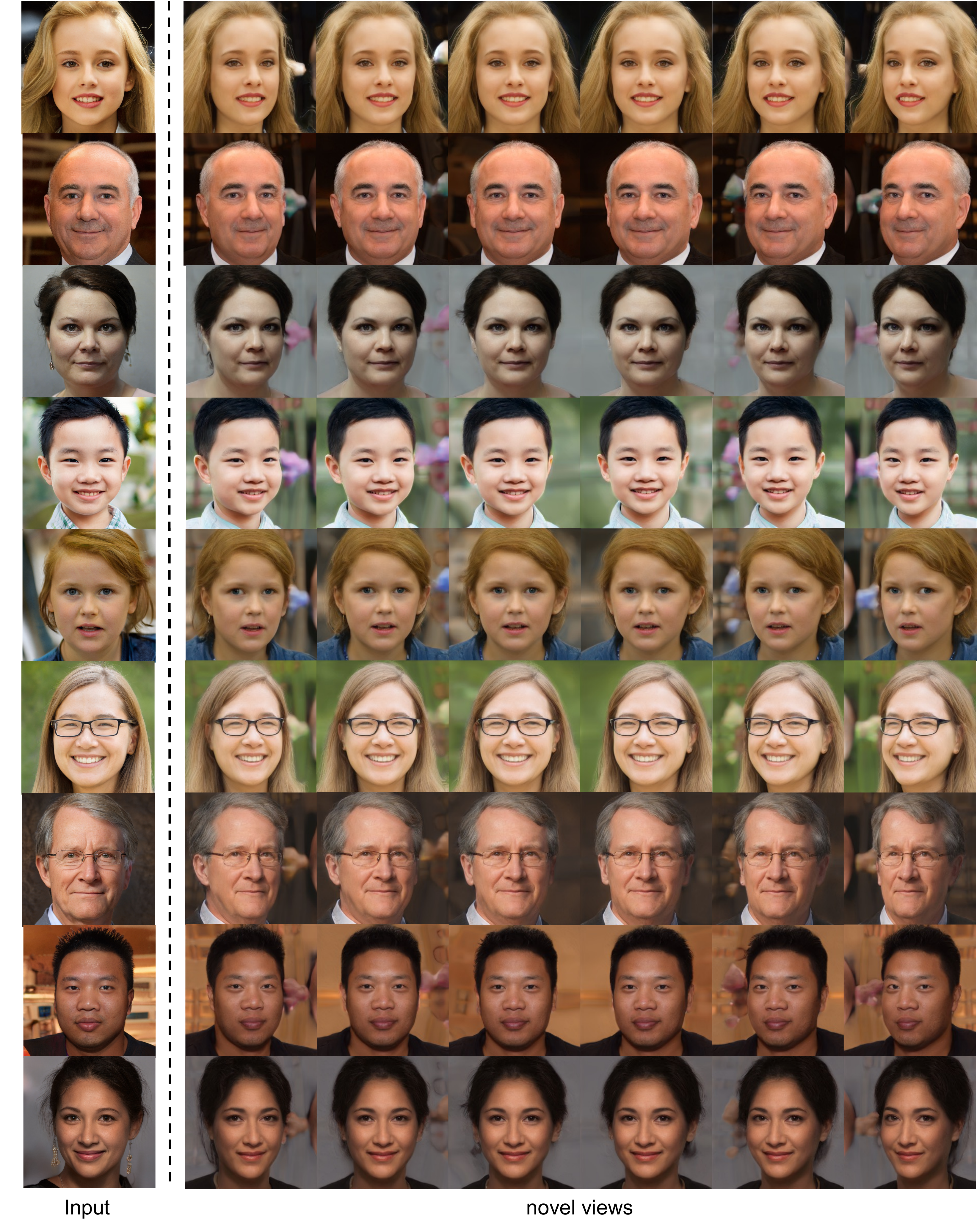}
  \caption{{More qualitative results using our encoder for novel view rendering on StyleGAN2-Fake using StyleNeRF generator.}}
  \label{fig:more_novel}
\end{figure*}

\paragraph{More qualitative results on novel rendering.}
We present more results on novel view rendering for different input images in Figure~\ref{fig:more_novel}. This figure demonstrates the generalization of our encoders plus the StyleNeRF generator to different races, gender, age, and skin tones. 

\begin{figure*}[t]
  \centering
  \includegraphics[width=0.9\linewidth]{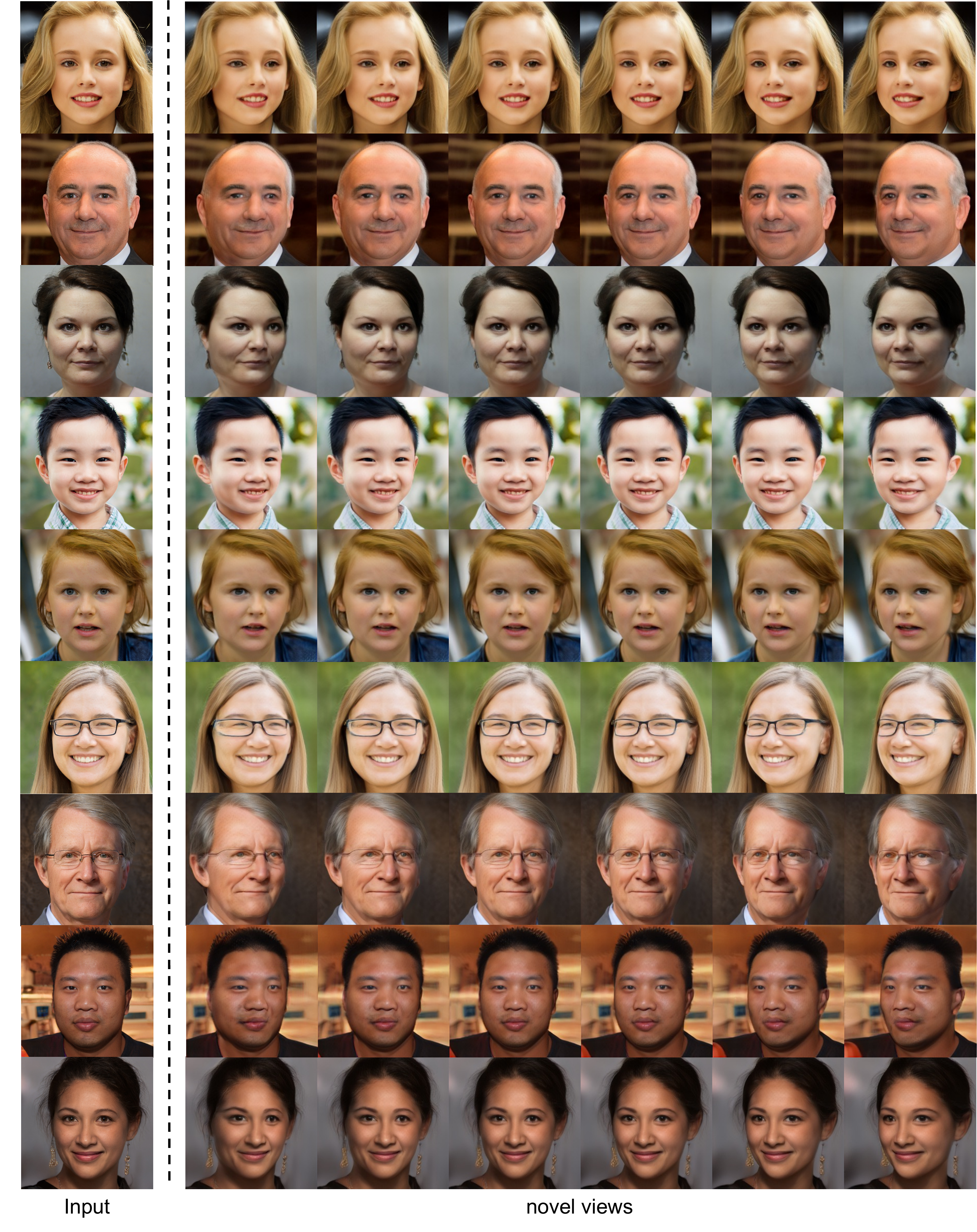}
  \caption{{Results generated by combining our encoder with PTI and \textbf{Eg3D} generator for novel view rendering on StyleGAN2-Fake}.}
  \label{fig:more_novel_eg3d}
\end{figure*}

\paragraph{Generator: StyleNeRF~\cite{gu2021stylenerf} vs. EG3D~\cite{chan2022efficient}}
To analyze the significance of the generators for the NeRF inversion, we also compare the results replacing the StyleNeRF generator with Eg3D using the same input image from Figure~\ref{fig:more_novel}. EG3D~\cite{chan2022efficient} is composed of StyleGAN2 architecture and utilizes tri-plane volume rendering. More details can be referred to in their paper. We re-train the encoders using the generator EG3D~\cite{chan2022efficient} and present the results of novel-view rendering in Figure~\ref{fig:more_novel_eg3d}.


\section{More experiments on animal faces}\label{app:more}

\begin{figure*}[t]
  \centering
  \includegraphics[width=0.9\linewidth]{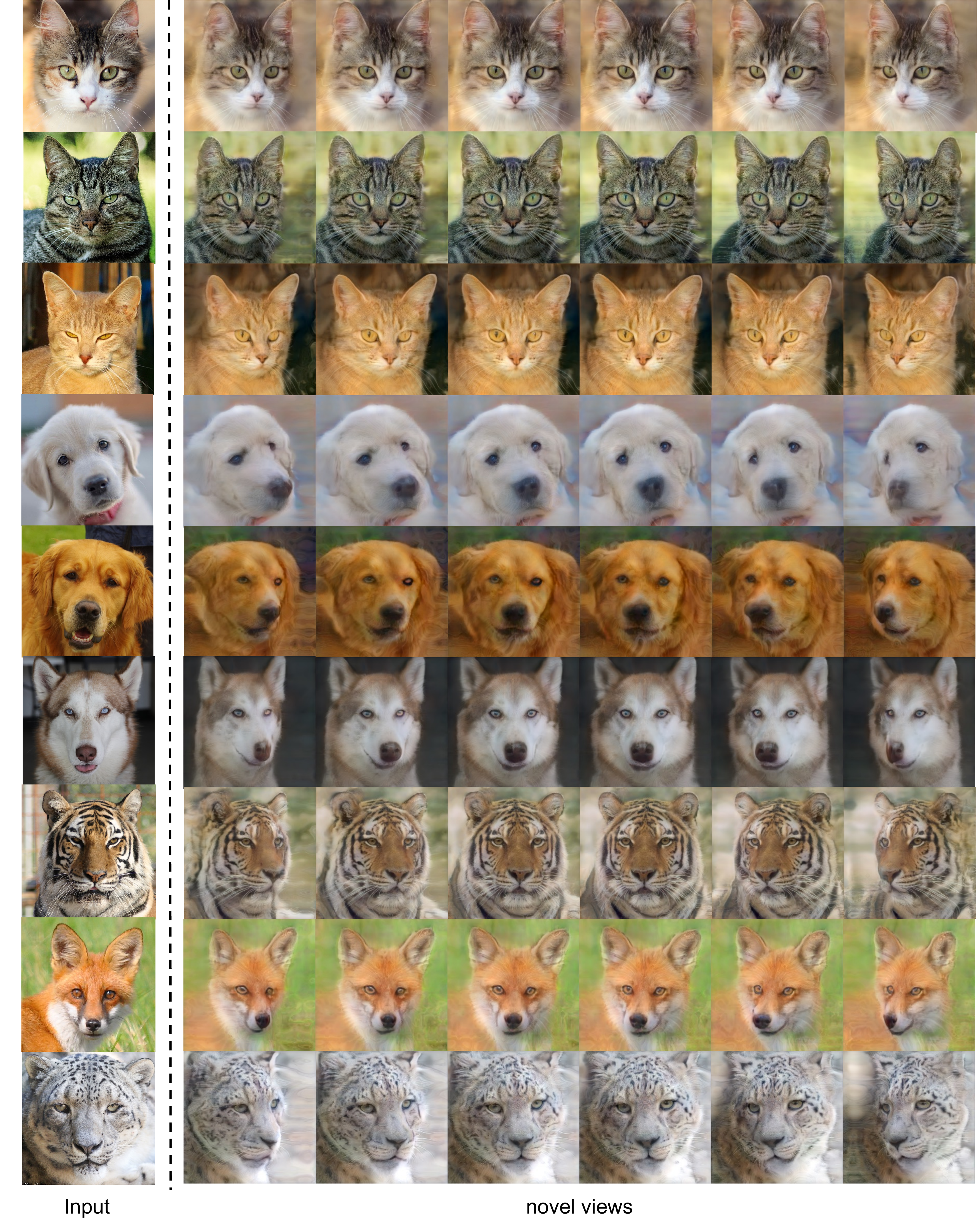}
  \caption{Additinal qualitative results using our encoder for novel view rendering on AFHQ. Note that since the checkpoint of StyleNeRF for AFHQ is not released, we train a sup-optimal checkpoint ourselves.}
  \label{fig:afhq_novel}
\end{figure*}

\begin{figure*}[t]
  \centering
  \includegraphics[width=0.9\linewidth]{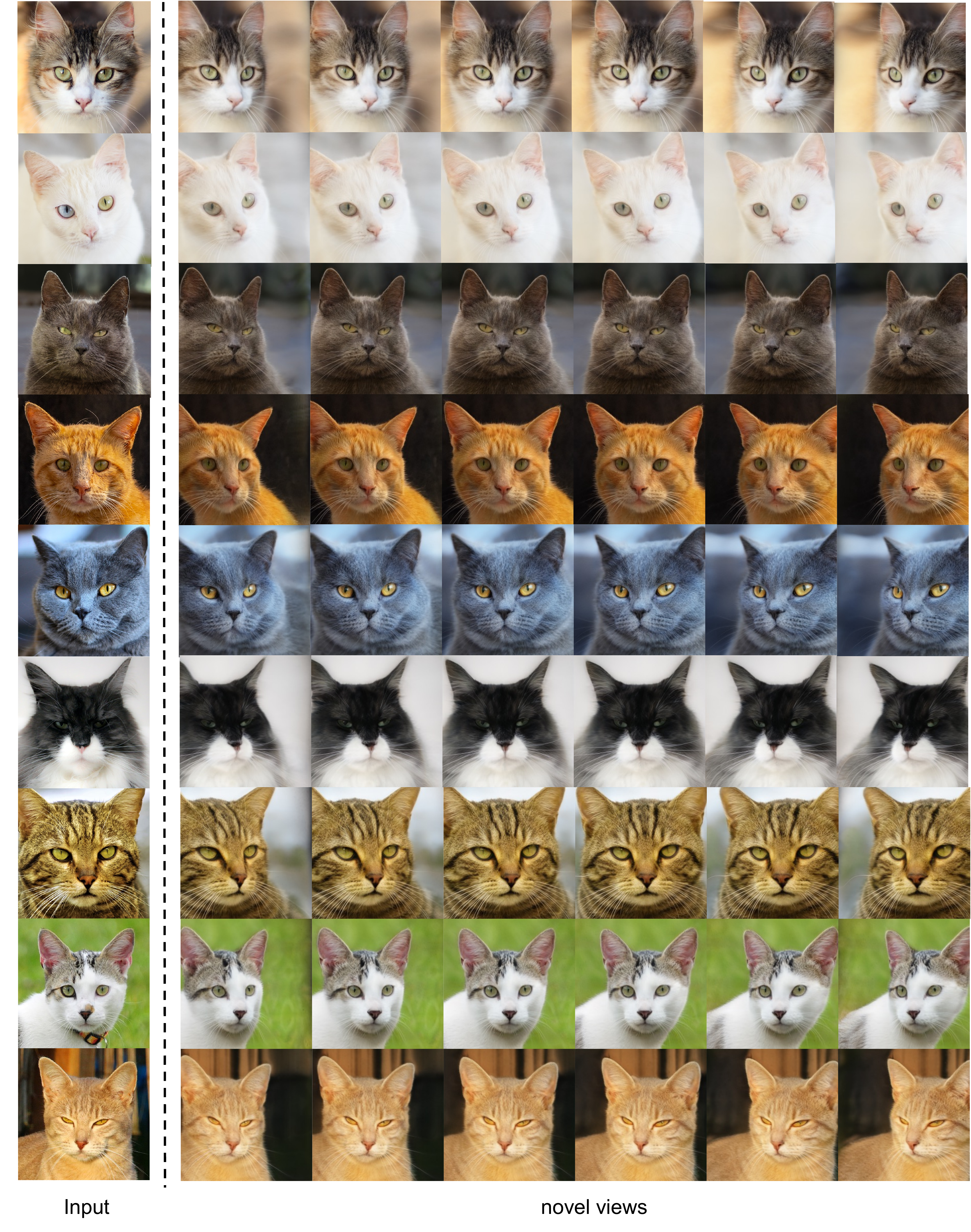}
  \caption{Additinal qualitative results using our encoder for novel view rendering on Cats subset in AFHQ using \textbf{Eg3D} as the generator ($\mathbf{G}$).}
  \label{fig:afhq_novel_eg3d}
\end{figure*}

To analyze the ability of our model on the inversion of animal faces, we conduct the experiments using AFHQ~\cite{choi2020starganv2}. This dataset includes three categories of animals which are cats, dogs, and wildlife. Since StyleNeRF~\cite{gu2021stylenerf} does not release the checkpoint for this dataset, we train our own checkpoint using the open-source code ourselves which may have sub-optimal rendering effectiveness. In addition, since we do not have a suitable off-the-shelf pose estimator for the animals, we additionally train the pose encoder in our base encoder (as shown in Figure~\ref{fig:encoder}) for estimating the camera pose for the animals. We present the results of NeRF inversion using our encoders in Figure~\ref{fig:afhq_novel}. In addition, we also present the NeRF inversion using the checkpoint of cat (a subset of AFHQ) released by EG3D~\cite{chan2022efficient} in Figure~\ref{fig:afhq_novel_eg3d}. These two figures show that our framework is able to perform effective 3D-aware NeRF inversion on animal faces.

\section{More experiments on stylization with CLIP}\label{app:clip}
\begin{figure*}[t]
  \centering
  \includegraphics[width=0.9\linewidth]{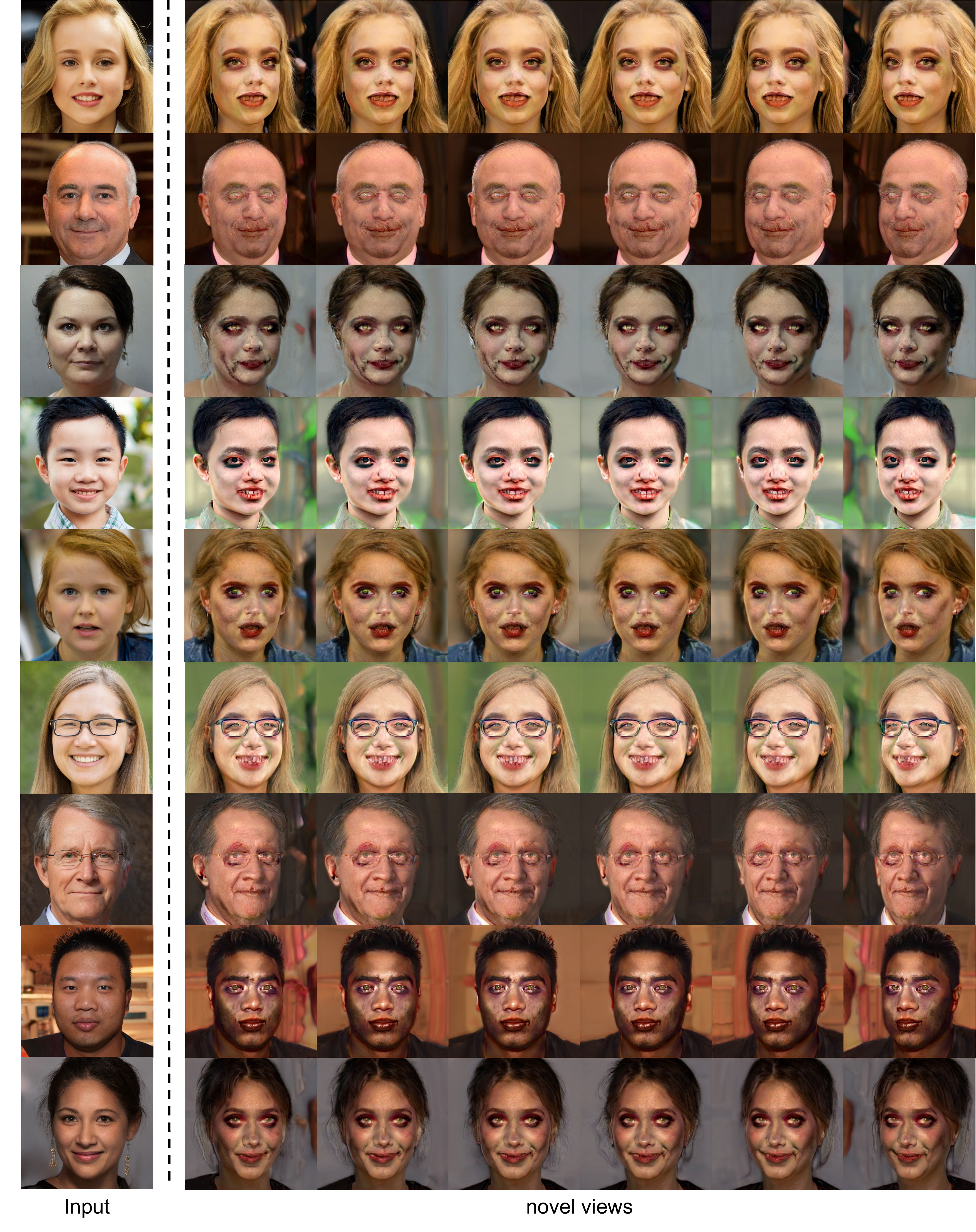}
  \caption{Stylization with the text "Zombie" using CLIP on the latent produced by our model. StyleNeRF is used as the generator.}
  \label{fig:zombie}
\end{figure*}

\begin{figure*}[t]
  \centering
  \includegraphics[width=0.9\linewidth]{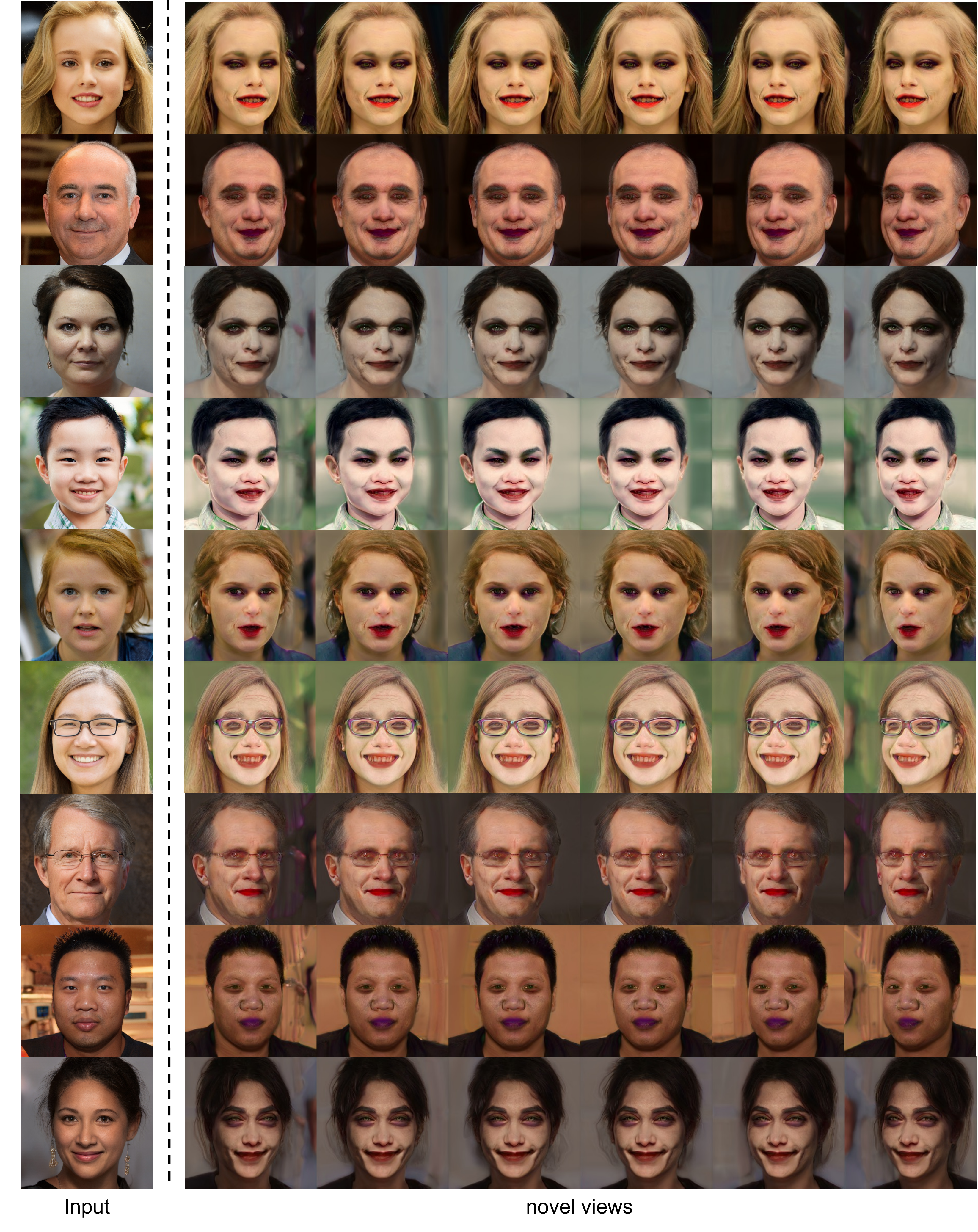}
  \caption{Stylization with the text "Joker" using CLIP on the latent produced by our model. StyleNeRF is used as the generator.}
  \label{fig:joker}
\end{figure*}

\begin{figure*}[t]
  \centering
  \includegraphics[width=0.9\linewidth]{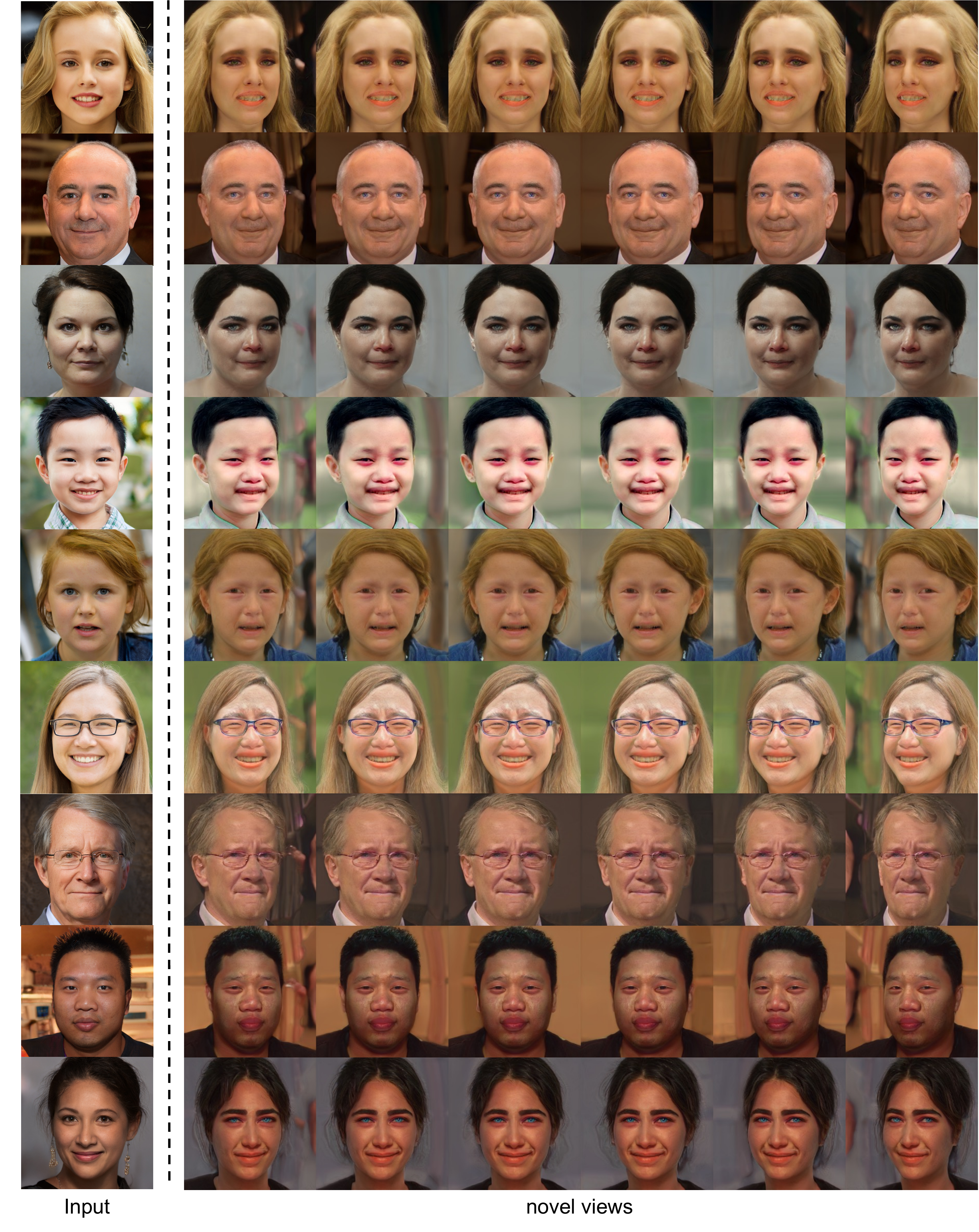}
  \caption{Stylization with the text "Crying face" using CLIP on the latent produced by our model. StyleNeRF is used as the generator.}
  \label{fig:cry}
\end{figure*}

\begin{figure*}[t]
  \centering
  \includegraphics[width=0.9\linewidth]{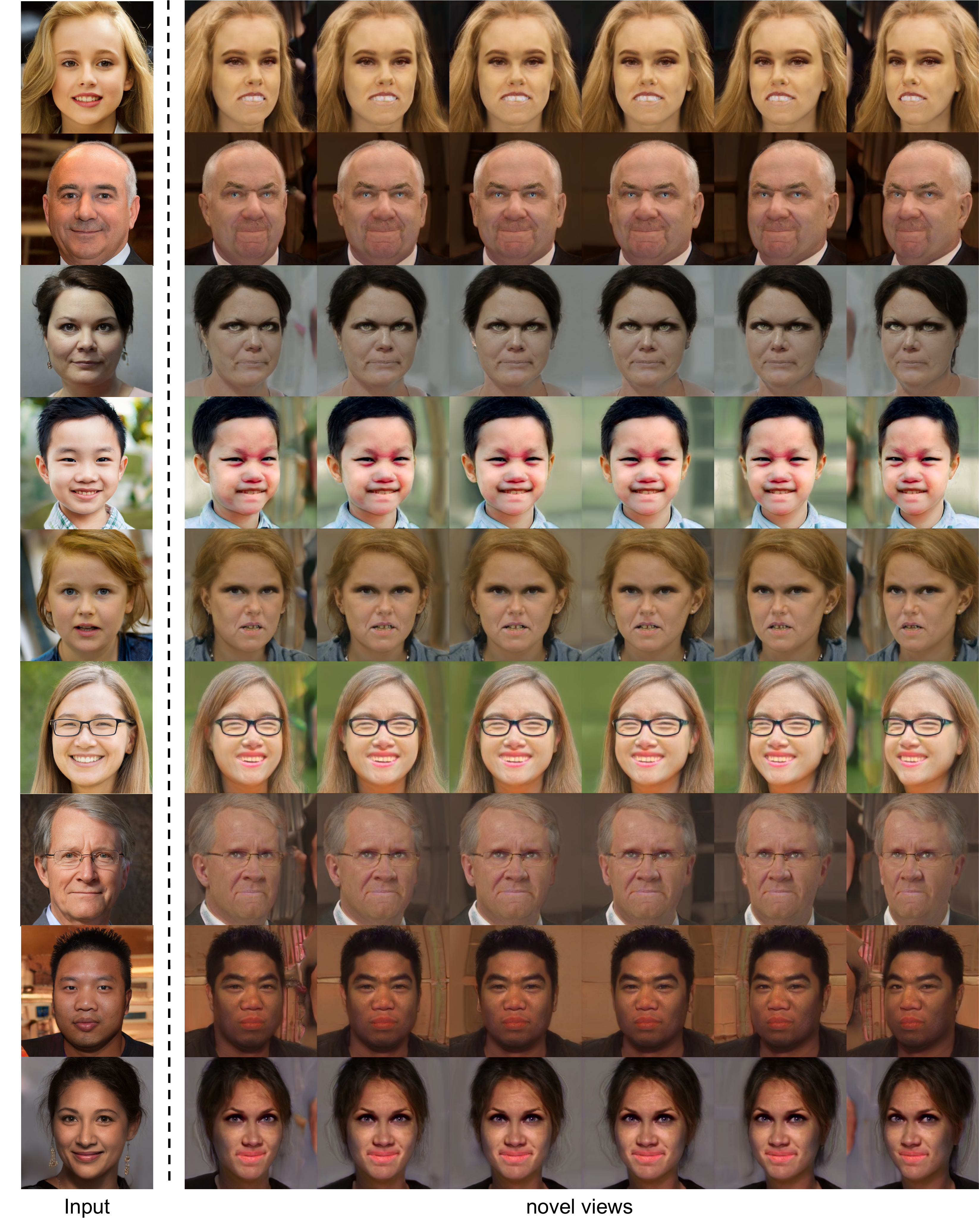}
  \caption{Stylization with the text "Angry face" using CLIP on the latent produced by our model. StyleNeRF is used as the generator.}
  \label{fig:angry}
\end{figure*}

\begin{figure*}[t]
  \centering
  \includegraphics[width=0.9\linewidth]{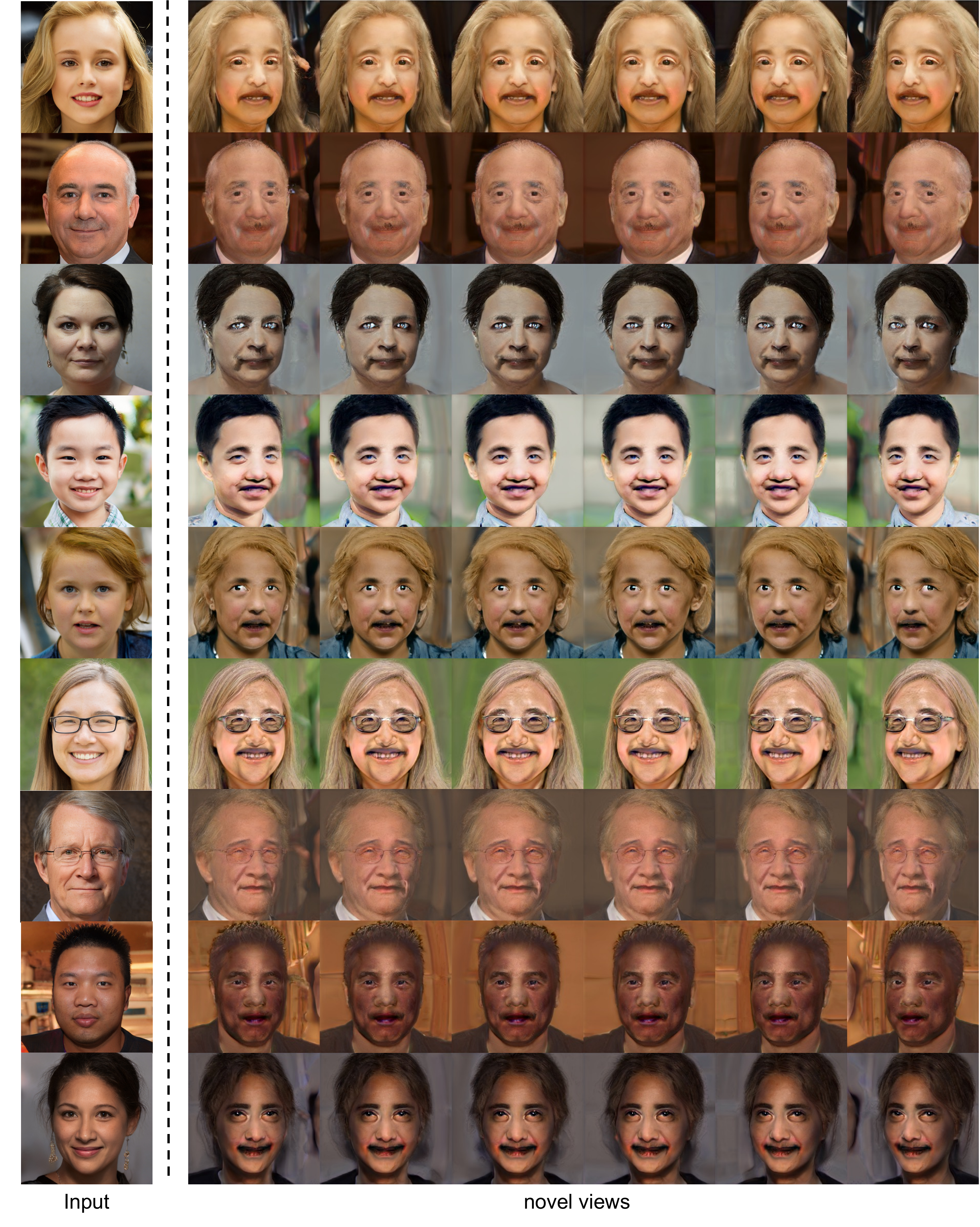}
  \caption{Stylization with the text "Albert Einstein" using CLIP on the latent produced by our model. StyleNeRF is used as the generator.}
  \label{fig:albert}
\end{figure*}

To demonstrate the generalization of our proposed 3D-aware encoder for stylization on the $W+$ space, we utilize CLIP~\cite{radford2021learning} to further edit the produced latent from our NeRF-3DE and present the results as follows:
\begin{itemize}
    \item Stylization with the text "Zombie" (Figure~\ref{fig:zombie}).
    \item Stylization with the text "Joker" (Figure~\ref{fig:joker}).
    \item Stylization with the text "Crying face" (Figure~\ref{fig:cry}).
    \item Stylization with the text "Angry face" (Figure~\ref{fig:angry}).
    \item Stylization with the text "Albert Einstein" (Figure~\ref{fig:albert}).
\end{itemize}

\section{Code}
We implement our model using PyTorch. The source code for this paper is provided in the directory named ``Code" for review. PLEASE DO NOT DISTRIBUTE ANY SOURCE FILES. The official version will be released after acceptance.

%% file: main.bbl
\begin{thebibliography}{10}\itemsep=-1pt

\bibitem{abdal2019image2stylegan}
Rameen Abdal, Yipeng Qin, and Peter Wonka.
\newblock Image2stylegan: How to embed images into the stylegan latent space?
\newblock In {\em CVPR}, pages 4432--4441, 2019.

\bibitem{abdal2020image2stylegan++}
Rameen Abdal, Yipeng Qin, and Peter Wonka.
\newblock Image2stylegan++: How to edit the embedded images?
\newblock In {\em CVPR}, pages 8296--8305, 2020.

\bibitem{abdal2021styleflow}
Rameen Abdal, Peihao Zhu, Niloy~J Mitra, and Peter Wonka.
\newblock Styleflow: Attribute-conditioned exploration of stylegan-generated
  images using conditional continuous normalizing flows.
\newblock {\em ACM Transactions on Graphics (ToG)}, 40(3):1--21, 2021.

\bibitem{alaluf2021restyle}
Yuval Alaluf, Or Patashnik, and Daniel Cohen-Or.
\newblock Restyle: A residual-based stylegan encoder via iterative refinement.
\newblock In {\em ICCV}, pages 6711--6720, 2021.

\bibitem{bau2020semantic}
David Bau, Hendrik Strobelt, William Peebles, Jonas Wulff, Bolei Zhou, Jun-Yan
  Zhu, and Antonio Torralba.
\newblock Semantic photo manipulation with a generative image prior.
\newblock {\em ACM Trans. Graph.}, 2020.

\bibitem{brock2018large}
Andrew Brock, Jeff Donahue, and Karen Simonyan.
\newblock Large scale gan training for high fidelity natural image synthesis.
\newblock In {\em ICLR}, 2018.

\bibitem{chan2022efficient}
Eric~R Chan, Connor~Z Lin, Matthew~A Chan, Koki Nagano, Boxiao Pan, Shalini
  De~Mello, Orazio Gallo, Leonidas~J Guibas, Jonathan Tremblay, Sameh Khamis,
  et~al.
\newblock Efficient geometry-aware 3d generative adversarial networks.
\newblock In {\em CVPR}, pages 16123--16133, 2022.

\bibitem{chan2021pi}
Eric~R Chan, Marco Monteiro, Petr Kellnhofer, Jiajun Wu, and Gordon Wetzstein.
\newblock pi-gan: Periodic implicit generative adversarial networks for
  3d-aware image synthesis.
\newblock In {\em CVPR}, pages 5799--5809, 2021.

\bibitem{chen2022sofgan}
Anpei Chen, Ruiyang Liu, Ling Xie, Zhang Chen, Hao Su, and Jingyi Yu.
\newblock Sofgan: A portrait image generator with dynamic styling.
\newblock {\em ACM Transactions on Graphics (TOG)}, 41(1):1--26, 2022.

\bibitem{chen2022sem2nerf}
Yuedong Chen, Qianyi Wu, Chuanxia Zheng, Tat-Jen Cham, and Jianfei Cai.
\newblock Sem2nerf: Converting single-view semantic masks to neural radiance
  fields.
\newblock {\em ECCV}, 2022.

\bibitem{choi2020starganv2}
Yunjey Choi, Youngjung Uh, Jaejun Yoo, and Jung-Woo Ha.
\newblock Stargan v2: Diverse image synthesis for multiple domains.
\newblock In {\em CVPR}, 2020.

\bibitem{deng2019arcface}
Jiankang Deng, Jia Guo, Niannan Xue, and Stefanos Zafeiriou.
\newblock Arcface: Additive angular margin loss for deep face recognition.
\newblock In {\em CVPR}, pages 4690--4699, 2019.

\bibitem{goetschalckx2019ganalyze}
Lore Goetschalckx, Alex Andonian, Aude Oliva, and Phillip Isola.
\newblock Ganalyze: Toward visual definitions of cognitive image properties.
\newblock In {\em ICCV}, pages 5744--5753, 2019.

\bibitem{goodfellow2014generative}
Ian Goodfellow, Jean Pouget-Abadie, Mehdi Mirza, Bing Xu, David Warde-Farley,
  Sherjil Ozair, Aaron Courville, and Yoshua Bengio.
\newblock Generative adversarial nets.
\newblock {\em Advances in neural information processing systems}, 27, 2014.

\bibitem{gu2021stylenerf}
Jiatao Gu, Lingjie Liu, Peng Wang, and Christian Theobalt.
\newblock Stylenerf: A style-based 3d-aware generator for high-resolution image
  synthesis.
\newblock {\em ICLR}, 2021.

\bibitem{gu2020image}
Jinjin Gu, Yujun Shen, and Bolei Zhou.
\newblock Image processing using multi-code gan prior.
\newblock In {\em Proceedings of the IEEE/CVF conference on computer vision and
  pattern recognition}, pages 3012--3021, 2020.

\bibitem{guan2020collaborative}
Shanyan Guan, Ying Tai, Bingbing Ni, Feida Zhu, Feiyue Huang, and Xiaokang
  Yang.
\newblock Collaborative learning for faster stylegan embedding.
\newblock {\em CORR}, 2020.

\bibitem{harkonen2020ganspace}
Erik H{\"a}rk{\"o}nen, Aaron Hertzmann, Jaakko Lehtinen, and Sylvain Paris.
\newblock Ganspace: Discovering interpretable gan controls.
\newblock {\em NeurIPS}, 33:9841--9850, 2020.

\bibitem{huang2020curricularface}
Yuge Huang, Yuhan Wang, Ying Tai, Xiaoming Liu, Pengcheng Shen, Shaoxin Li,
  Jilin Li, and Feiyue Huang.
\newblock Curricularface: adaptive curriculum learning loss for deep face
  recognition.
\newblock In {\em CVPR}, pages 5901--5910, 2020.

\bibitem{jahanian2019steerability}
Ali Jahanian, Lucy Chai, and Phillip Isola.
\newblock On the" steerability" of generative adversarial networks.
\newblock {\em ICLR}, 2020.

\bibitem{kajiya1984ray}
James~T Kajiya and Brian~P Von~Herzen.
\newblock Ray tracing volume densities.
\newblock {\em ACM SIGGRAPH computer graphics}, 18(3):165--174, 1984.

\bibitem{kang2021gan}
Kyoungkook Kang, Seongtae Kim, and Sunghyun Cho.
\newblock Gan inversion for out-of-range images with geometric transformations.
\newblock In {\em ICCV}, pages 13941--13949, 2021.

\bibitem{karras2017progressive}
Tero Karras, Timo Aila, Samuli Laine, and Jaakko Lehtinen.
\newblock Progressive growing of gans for improved quality, stability, and
  variation.
\newblock In {\em ICLR}, 2018.

\bibitem{karras2019style}
Tero Karras, Samuli Laine, and Timo Aila.
\newblock A style-based generator architecture for generative adversarial
  networks.
\newblock In {\em CVPR}, pages 4401--4410, 2019.

\bibitem{Karras2019stylegan2}
Tero Karras, Samuli Laine, Miika Aittala, Janne Hellsten, Jaakko Lehtinen, and
  Timo Aila.
\newblock Analyzing and improving the image quality of {StyleGAN}.
\newblock In {\em CVPR}, 2020.

\bibitem{kim2021exploiting}
Hyunsu Kim, Yunjey Choi, Junho Kim, Sungjoo Yoo, and Youngjung Uh.
\newblock Exploiting spatial dimensions of latent in gan for real-time image
  editing.
\newblock In {\em CVPR}, pages 852--861, 2021.

\bibitem{liu2015deep}
Ziwei Liu, Ping Luo, Xiaogang Wang, and Xiaoou Tang.
\newblock Deep learning face attributes in the wild.
\newblock In {\em ICCV}, pages 3730--3738, 2015.

\bibitem{mildenhall2020nerf}
Ben Mildenhall, Pratul~P Srinivasan, Matthew Tancik, Jonathan~T Barron, Ravi
  Ramamoorthi, and Ren Ng.
\newblock Nerf: Representing scenes as neural radiance fields for view
  synthesis.
\newblock In {\em ECCV}, pages 405--421. Springer, 2020.

\bibitem{niemeyer2021giraffe}
Michael Niemeyer and Andreas Geiger.
\newblock Giraffe: Representing scenes as compositional generative neural
  feature fields.
\newblock In {\em CVPR}, pages 11453--11464, 2021.

\bibitem{or2022stylesdf}
Roy Or-El, Xuan Luo, Mengyi Shan, Eli Shechtman, Jeong~Joon Park, and Ira
  Kemelmacher-Shlizerman.
\newblock Stylesdf: High-resolution 3d-consistent image and geometry
  generation.
\newblock In {\em CVPR}, pages 13503--13513, 2022.

\bibitem{pan2021shading}
Xingang Pan, Xudong Xu, Chen~Change Loy, Christian Theobalt, and Bo Dai.
\newblock A shading-guided generative implicit model for shape-accurate
  3d-aware image synthesis.
\newblock {\em NeurIPS}, 34:20002--20013, 2021.

\bibitem{pidhorskyi2020adversarial}
Stanislav Pidhorskyi, Donald~A Adjeroh, and Gianfranco Doretto.
\newblock Adversarial latent autoencoders.
\newblock In {\em CVPR}, pages 14104--14113, 2020.

\bibitem{radford2021learning}
Alec Radford, Jong~Wook Kim, Chris Hallacy, Aditya Ramesh, Gabriel Goh,
  Sandhini Agarwal, Girish Sastry, Amanda Askell, Pamela Mishkin, Jack Clark,
  et~al.
\newblock Learning transferable visual models from natural language
  supervision.
\newblock In {\em International Conference on Machine Learning}, pages
  8748--8763. PMLR, 2021.

\bibitem{richardson2021encoding}
Elad Richardson, Yuval Alaluf, Or Patashnik, Yotam Nitzan, Yaniv Azar, Stav
  Shapiro, and Daniel Cohen-Or.
\newblock Encoding in style: a stylegan encoder for image-to-image translation.
\newblock In {\em CVPR}, pages 2287--2296, 2021.

\bibitem{roich2021pivotal}
Daniel Roich, Ron Mokady, Amit~H Bermano, and Daniel Cohen-Or.
\newblock Pivotal tuning for latent-based editing of real images.
\newblock {\em arXiv preprint arXiv:2106.05744}, 2021.

\bibitem{ruiz2018fine}
Nataniel Ruiz, Eunji Chong, and James~M Rehg.
\newblock Fine-grained head pose estimation without keypoints.
\newblock In {\em Proceedings of the IEEE conference on computer vision and
  pattern recognition workshops}, pages 2074--2083, 2018.

\bibitem{schwarz2020graf}
Katja Schwarz, Yiyi Liao, Michael Niemeyer, and Andreas Geiger.
\newblock Graf: Generative radiance fields for 3d-aware image synthesis.
\newblock {\em NeurIPS}, 33:20154--20166, 2020.

\bibitem{shen2020interpreting}
Yujun Shen, Jinjin Gu, Xiaoou Tang, and Bolei Zhou.
\newblock Interpreting the latent space of gans for semantic face editing.
\newblock In {\em CVPR}, pages 9243--9252, 2020.

\bibitem{shoshan2021gan}
Alon Shoshan, Nadav Bhonker, Igor Kviatkovsky, and Gerard Medioni.
\newblock Gan-control: Explicitly controllable gans.
\newblock In {\em ICCV}, pages 14083--14093, 2021.

\bibitem{sun2022ide}
Jingxiang Sun, Xuan Wang, Yichun Shi, Lizhen Wang, Jue Wang, and Yebin Liu.
\newblock Ide-3d: Interactive disentangled editing for high-resolution 3d-aware
  portrait synthesis.
\newblock {\em Siggraph Asia}, 2022.

\bibitem{tewari2020stylerig}
Ayush Tewari, Mohamed Elgharib, Gaurav Bharaj, Florian Bernard, Hans-Peter
  Seidel, Patrick P{\'e}rez, Michael Zollhofer, and Christian Theobalt.
\newblock Stylerig: Rigging stylegan for 3d control over portrait images.
\newblock In {\em CVPR}, pages 6142--6151, 2020.

\bibitem{tov2021designing}
Omer Tov, Yuval Alaluf, Yotam Nitzan, Or Patashnik, and Daniel Cohen-Or.
\newblock Designing an encoder for stylegan image manipulation.
\newblock {\em ACM Transactions on Graphics (TOG)}, 40(4):1--14, 2021.

\bibitem{voynov2020unsupervised}
Andrey Voynov and Artem Babenko.
\newblock Unsupervised discovery of interpretable directions in the gan latent
  space.
\newblock In {\em Proceedings of the International Conference on Machine
  Learning (ICML)}, pages 9786--9796. PMLR, 2020.

\bibitem{wang2021geometry}
Binxu Wang and Carlos~R Ponce.
\newblock The geometry of deep generative image models and its applications.
\newblock In {\em ICLR}, 2021.

\bibitem{wang2022high}
Tengfei Wang, Yong Zhang, Yanbo Fan, Jue Wang, and Qifeng Chen.
\newblock High-fidelity gan inversion for image attribute editing.
\newblock In {\em CVPR}, pages 11379--11388, 2022.

\bibitem{xu20223d}
Yinghao Xu, Sida Peng, Ceyuan Yang, Yujun Shen, and Bolei Zhou.
\newblock 3d-aware image synthesis via learning structural and textural
  representations.
\newblock In {\em CVPR}, pages 18430--18439, 2022.

\bibitem{zhang2019self}
Han Zhang, Ian Goodfellow, Dimitris Metaxas, and Augustus Odena.
\newblock Self-attention generative adversarial networks.
\newblock In {\em Proceedings of the International Conference on Machine
  Learning (ICML)}, pages 7354--7363. PMLR, 2019.

\bibitem{zhang2018unreasonable}
Richard Zhang, Phillip Isola, Alexei~A Efros, Eli Shechtman, and Oliver Wang.
\newblock The unreasonable effectiveness of deep features as a perceptual
  metric.
\newblock In {\em CVPR}, pages 586--595, 2018.

\bibitem{zhao2022generative}
Xiaoming Zhao, Fangchang Ma, David G{\"u}era, Zhile Ren, Alexander~G Schwing,
  and Alex Colburn.
\newblock Generative multiplane images: Making a 2d gan 3d-aware.
\newblock In {\em ECCV}, 2022.

\bibitem{zhu2020indomain}
Jiapeng Zhu, Yujun Shen, Deli Zhao, and Bolei Zhou.
\newblock In-domain gan inversion for real image editing.
\newblock In {\em ECCV}, 2020.

\bibitem{zhu2016generative}
Jun-Yan Zhu, Philipp Kr{\"a}henb{\"u}hl, Eli Shechtman, and Alexei~A Efros.
\newblock Generative visual manipulation on the natural image manifold.
\newblock In {\em ECCV}, pages 597--613. Springer, 2016.

\end{thebibliography}
